\documentclass[10pt,twocolumn,letterpaper]{article}

\usepackage[pagenumbers]{cvpr} 

\renewcommand{\paragraph}[1]{\vspace{.5em}\noindent\textbf{#1.}}

\definecolor{cvprblue}{rgb}{0.21,0.49,0.74}
\usepackage[pagebackref,breaklinks,colorlinks,allcolors=cvprblue]{hyperref}
\usepackage{pifont}
\usepackage{overpic}
\usepackage{caption}
\usepackage[table]{xcolor}
\usepackage{colortbl}
\definecolor{mygray}{gray}{.94}
\usepackage{amsmath}
\usepackage{amssymb}
\usepackage{booktabs}
\usepackage{makecell}
\usepackage[accsupp]{axessibility}

\def\paperID{2306} 
\def\confName{CVPR}
\def\confYear{2026}

\title{Gloria: Consistent Character Video Generation via Content Anchors}

\author{Yuhang Yang$^{1}$, Fan Zhang$^{2}$, Huaijin Pi$^{3}$, Shuai Guo$^{1}$, Guowei Xu$^{4}$, Wei Zhai$^{1,\dagger}$ \\ Yang Cao$^{1}$, Zheng-Jun Zha$^{1}$\\
$^{1}$~MoE Key Laboratory of Brain-inspired Intelligent Perception and Cognition, USTC \\ $^{2}$~UNSW \textit{}  $^{3}$~HKU \textit{} $^{4}$~UESTC\\
$\dagger$Corresponding Author,  \href{https://yyvhang.github.io/Gloria_Page/}{https://yyvhang.github.io/Gloria\_Page/}
}

\begin{document}
\twocolumn[{%
         \renewcommand\twocolumn[1][]{#1}%
         \maketitle
         \begin{center}
            \centering
            \includegraphics[width=0.95\textwidth]{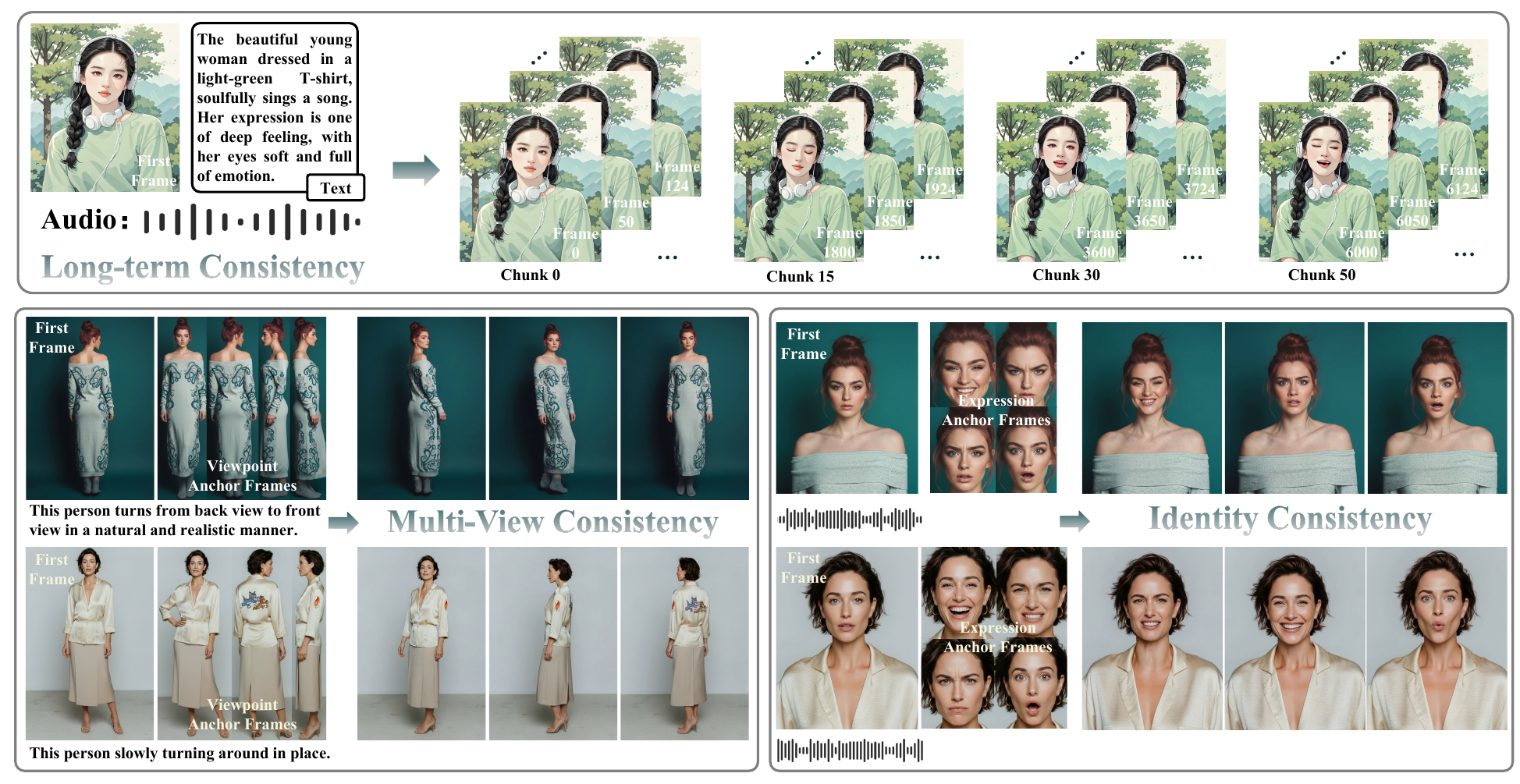}
            \captionof{figure}{Gloria achieves long-term, multi-view appearance and expressive identity consistency of characters in video generation through character-centric anchor frames, supporting text, images, and audio as inputs.}
            \label{fig:teaser}
         \end{center}
}]
\begin{abstract}
Digital characters are central to modern media, yet generating character videos with long-duration, consistent multi-view appearance and expressive identity remains challenging. Existing approaches either provide insufficient context to preserve identity or leverage non-character-centric information as the ``memory", leading to suboptimal consistency.
Recognizing that character video generation inherently resembles an ``outside-looking-in" scenario. In this work, we propose representing the character’s visual attributes through a compact set of anchor frames.
This design provides stable references for consistency, while reference-based video generation inherently faces challenges of copy-pasting and multi-reference conflicts. To address these, we introduce two mechanisms: Superset Content Anchoring, providing intra- and extra-training clip cues to prevent duplication, and RoPE as Weak Condition, encoding positional offsets to distinguish multiple anchors.
Furthermore, we construct a scalable pipeline to extract these anchors from massive videos. Experiments show our method generates high-quality character videos exceeding $10$ minutes, and achieves expressive identity and appearance consistency across views, surpassing existing methods.
\end{abstract}
    
\section{Introduction}
\label{sec:intro}

Digital characters serve as the primary medium for user interaction in applications ranging from films, short-form videos to immersive games. Recently, large-scale diffusion-based video generation models \cite{wan2025,polyak2024movie,brooks2024video,wiedemer2025video,kong2024hunyuanvideo,kuaishou2024kling,hacohen2024ltx} greatly improves the fidelity of dynamics and appearance of digital characters, the video-based synthesis of vivid characters has emerged as a highly promising direction.

Despite recent progress, generating long-duration, multi-view and expressive identity (ID) consist character videos still remains highly challenging. Input formats of existing methods \cite{fei2025skyreels,lin2025omnihuman1,jiang2025omnihuman,meng2025echomimicv3,meng2025echomimicv3,hu2025HunyuanVideo-Avatar,wang2025fantasytalking,wei2025mocha}, such as a single reference image or a text prompt provide insufficient context to preserve a character’s visual attributes across viewpoints and expressions. As the generated sequence grows longer, error accumulation in DiT-based video generation further degrades appearance fidelity and identity consistency. To mitigate these issues, several methods \cite{yang2025infinitetalkaudiodrivenvideogeneration,gao2025wans2vaudiodrivencinematicvideo,ding2025kling,gan2025omniavatar} incorporate pre-specified frames or generated frames as contextual ``memory." However, these frames are typically non-character-centric and ad-hoc picks, lacking stable and semantically grounded cues about the character’s multi-view appearance and expressive identity. As a result, the injected memory frames often provides only partial identity cues, ultimately leading to inconsistent character generation.

In this paper, we present a video-based solution for generating consistent long-duration character videos. Our key insight is to reframe character-centric video generation as an ``outside-looking-in" scenario: a character’s visual attributes such as multi-view appearance and expressive identity (Fig.~\ref{fig:teaser}) can be compactly represented by a structured set of anchor frames, termed \textbf{content anchors}. These anchors provide stable visual references, while the character’s diverse motions can be learned from short video clips.
During inference, an artist supplies a set of content anchors and instructions (\eg, audio or text), and the model generates long videos while preserving the character's appearance and identity. The anchor set includes frames capturing the overall scene, multiple viewpoints of the character, and faces with various expressions.
Although conceptually simple, implementing it still has several technical challenges:
1) how to inject anchor frames so that the model learns to associate relevant information rather than simply copy-pasting;
2) how to avoid conflicts when multiple anchors are injected simultaneously;
and 3) how to efficiently select corresponding anchor frames from a large data pool.

To tackle these challenges, we propose a solution that spans both modeling and data. We adopt a unified content-anchor injection mechanism that concatenates anchor tokens with video tokens, enabling the anchors to directly participate in self-attention and provide a stable reference. To avoid the copy-paste pattern, we introduce the ``Superset Content Anchoring'', which supplies each training clip with both intra- and extra- anchor frames, encouraging the model to adaptively extract relevant information rather than duplicating pixels. To avoid conflicts among multiple anchors, we introduce the ``RoPE \cite{su2024roformer} as Weak Condition'', which shifts different anchors to distinct positional ranges so the model can reliably differentiate them under mixed-ratio training. Moreover, we build an automated data pipeline that extracts anchor frames at scale. We identify viewpoint anchors by analyzing the subject’s facing direction relative to the camera \cite{24siga_gvhmr,15tog_smpl}, and extract expression anchors via emotion recognition \cite{savchenko2023facial,savchenko2022classifying} refined by MLLMs \cite{comanici2025gemini}, enabling high-quality, large-scale anchor construction.

Combining these modeling mechanisms with our scalable data pipeline, we ultimately enable a model that generates character videos over $10$ minutes and maintains multi-view appearance and expressive identity consistency. The contributions are summarized as follows:
\begin{itemize}[leftmargin=15pt,topsep=0pt,itemsep=2pt]
    \item[\textbf{1)}] We propose content anchors for character video generation, a compact and structured set of frames that serve as persistent visual references for long-term, multi-view appearance and expressive identity consistency.
    \item[\textbf{2)}] We introduce ``Superset Content Anchoring" and ``RoPE as weak condition" to eliminate the copy-paste pattern and multi-anchor conflicts, enabling unified training with multiple content anchors.
    \item[\textbf{3)}] We construct an automated data pipeline to extract anchor frames \eg, viewpoints and expressions. The experimental results show that our model outperforms existing methods in terms of long-term, appearance and ID consistency in character video generation.
\end{itemize}

\section{Related Work}
\label{sec:related}
\subsection{Condition-Driven Human Video Generation}

With the recent progress of video diffusion models \cite{wan2025,polyak2024movie,wiedemer2025video,kong2024hunyuanvideo,kuaishou2024kling,HaCohen2024LTXVideo,liu2025worldweaver,liu2025tuna}, human-centric visual generation \cite{23_animateanyone,liu2023cones,liu2023customizable,chen2024livephoto,hu2025polyvivid,liang2025omniv2v,huomnicustom,chen2025instancev} has emerged as an important field that aims to synthesize digital characters that closely approximate real human behaviors. Compared to 3D methods \cite{yang2025sigman,qiu2025lhm, yu2025hero,liu2025manganinja,hu2024motionmaster,shao2025great,han2025touch,liu2024grounding}, this paradigm is more promising. Building upon the foundation models, methods introduce various control modalities to enable more precise and expressive character animation. 
Among them, pose-driven approaches \cite{23_animateanyone,24cvpr_magicanimate,25iccv_vace,25_animateanyone2,li2024dispose,chen2025dancetogether,yang2023effective,wang2025poseanything} leverage 2D skeletons or keypoints to guide the motion trajectory, 
achieving spatially aligned and temporally coherent human movements. 
Meanwhile, audio-driven methods \cite{lin2025omnihuman1,hu2025HunyuanVideo-Avatar,wang2025fantasytalking,25aaai_echomimic,gan2025omniavatar, ding2025kling,kong2025let,gao2025wans2vaudiodrivencinematicvideo,wei2025mocha,fei2025skyreels} 
enable synchronized lip motion and facial expressions by coupling speech and visual dynamics.
Several recent studies explore long-duration human video generation \cite{jiang2025omnihuman,gao2025wans2vaudiodrivencinematicvideo,yang2025infinitetalkaudiodrivenvideogeneration}, typically extending temporal coherence from a few seconds to around one minute. 
Among them, WanS2V \cite{gao2025wans2vaudiodrivencinematicvideo} leverages the FramePack mechanism \cite{zhang2025packing} to aggregate multiple reference frames of the same person to improve temporal and appearance consistency, while other works \cite{ding2025kling,yang2025infinitetalkaudiodrivenvideogeneration} adopt keyframe-interpolation strategies to maintain stable identity over extended sequences. 
However, generating long-duration character videos is still far from satisfactory \cite{zeng2024dawn,yang2025videogen,ling2025vmbench,zhang2024evaluationagent}. 
Our approach aims to address these problems by introducing character-centric and compact \textit{content anchors}, providing  a persistent global reference and enabling long-duration consistent character video generation.

\subsection{Consistent Human Video Generation}
Maintaining consistent identity and appearance of a character is a key aspect in character video generation.
At the image level, reference-based diffusion models \cite{gpt4o,wu2025qwenimagetechnicalreport,seedream2025seedream,google_gemini_flash_image} achieve strong subject fidelity and 3D consistency through shared latent embeddings or reference-image conditioning.
However, extending this consistency to videos requires preserving a character’s geometry and appearance across diverse poses and viewpoints.
Some methods \cite{hogue2024diffted,lin2024cyberhost} enhance temporal and appearance coherence by incorporating explicit human priors such as skeletons or regional constraints to guide motion and stabilize identity.
Subsequent approaches \cite{lin2025omnihuman1,jiang2025omnihuman,hu2025HunyuanVideo-Avatar} shift toward implicit modeling, injecting reference-image information directly into the latent space and relying on large-scale training data to learn identity consistency.
More recent studies \cite{25_univideo,25_humo,liu2025phantom,chen2025multi,hu2025hunyuancustom,bao2024vidu,kuaishou2024kling} further extend this paradigm to multi-subject scenarios, employing full-attention mechanisms that represent all reference images as tokens for cross-subject appearance alignment.
In contrast, we focus on single-subject consistency and define a compact anchor set that includes multiple viewpoints and expressions of a character, offering persistent global guidance for stable identity and appearance.
  
\section{Data Pipeline}
\begin{figure}
    \centering
    \includegraphics[width=1.0\linewidth]{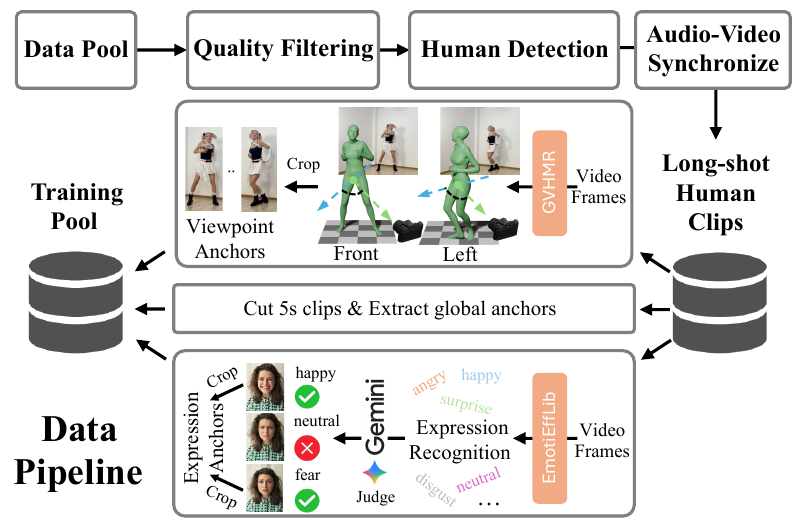}
    \caption{The pipeline to construct training clips and character-centric anchor frames, \eg, global, viewpoint, and expression. The blue arrow marks the subject’s forward orientation, whereas the green arrow marks the camera-facing direction.}
    \label{fig:data}
\end{figure}

The overall data pipeline is shown in Fig. \ref{fig:data}. Our goal is to obtain clean human-centric data and extract anchor frames covering the overall scene, four viewpoints (front, back, left, right), and eight facial expressions (surprise, angry, disgust, fear, contempt, sad, neutral, happy). Given a data pool, we refer pipelines \cite{polyak2024movie,wan2025} to filter low quality videos. Next, we apply human detection \cite{jiang2023rtmpose,wang2024yolov9} to keep only single-person clips and discard non-human or multi-human clips. After that, we further remove videos with audio-visual desynchronization \cite{Chung16a}. Eventually, we obtain long-shot single-person videos, which are then segmented into $5$s ($24$ fps) for training.

In addition, we construct a pipeline to extract character-centric anchor frames. For the global anchor frame, we randomly sample $10$ candidate frames from each long-shot video, including those within and beyond the training clips. 
Regarding the viewpoint anchors, we use GVHMR \cite{24siga_gvhmr} to estimate the human pose in the camera coordinate system and compute the angle between the human forward direction and the camera direction. And use thresholds to distinguish each viewpoint and extract corresponding anchor frames. 
For the expression anchor frame, we select source videos whose native resolution $\geq$ 1080p to obtain clear faces. Next, the EmotiEffLib \cite{savchenko2023facial,savchenko2022classifying} is leveraged to extract video clips exhibiting more than two of the $8$ predefined expressions. Followed by the Gemini \cite{comanici2025gemini} to judge whether each anchor frame matches its semantic expression and filter out wrong frames. On our constructed test set ($100$ viewpoint and $100$ expression samples), our data pipeline achieves $98\%$ accuracy in viewpoint-anchor extraction. Using EmotiEffLib alone yields $66\%$ accuracy for extracting expression anchor frames, while incorporating the Gemini judge improves the accuracy to $82\%$. Finally, for viewpoint and expression anchor frames, we respectively crop the human body and head to isolate the relevant features and prevent them from being coupled with extraneous information. 

\section{Method}
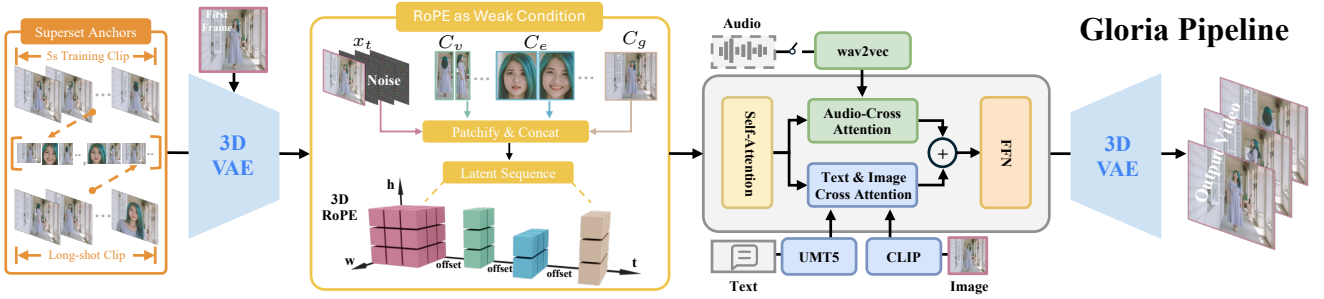
\begin{figure*}[t]
    \centering
    \begin{overpic}[width=1.0\linewidth]{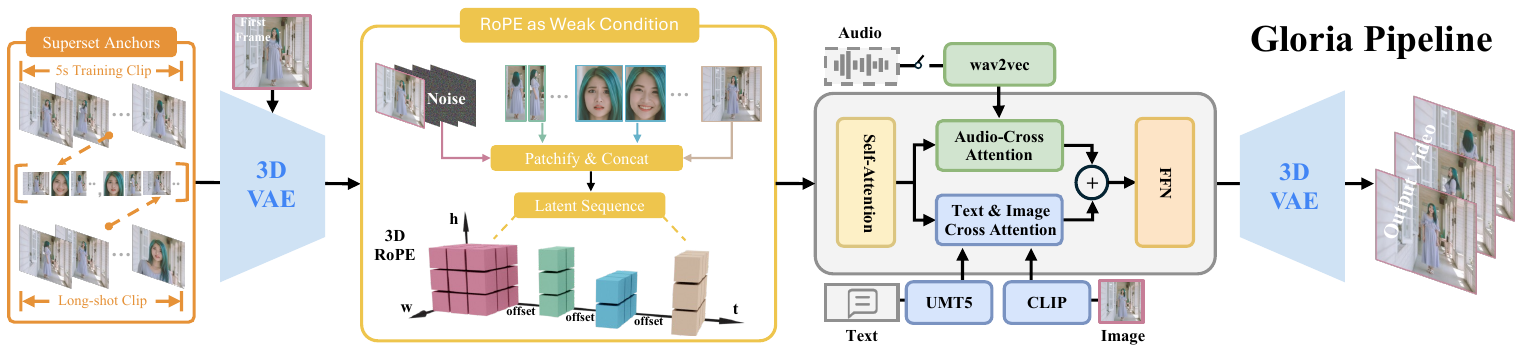}
    \put(27,19){{\textbf{\scriptsize $x_t$}}}
    \put(33.5,18.9){{\textbf{\scriptsize $C_v$}}}
    \put(40,18.9){{\textbf{\scriptsize $C_e$}}}
    \put(47.5,19.2){{\textbf{\scriptsize $C_g$}}}
    \end{overpic}
    \caption{Overall of the Gloria pipeline, which includes the source of content anchors (Superset Anchors), the manner to inject these anchors (RoPE as Weak Condition), and the overall framework with multi-modal conditions \eg, text and audio.}
    \label{fig:method}
\end{figure*}

Given multi-modal inputs including text, an image, optional audio, and a set of anchor frames, our goal is to generate long-term character videos that maintain appearance and identity consistency across multiple viewpoints and expressions.
Based on the Wan-I2V \cite{wan2025} architecture, we present Gloria, a unified framework for consistent character generation. The overall pipeline is illustrated in Fig. \ref{fig:method}.
We first introduce the design and injection mechanism of content anchors that provide structured reference frames for maintaining consistency (Sec. \ref{sec:method:anchor}).
Then we describe the manner to integrate control conditions \eg, text and audio to drive synchronized and expressive motion generation (Sec. \ref{sec:method:audio}).
Finally, we specify the inference process (Sec. \ref{sec:method:infer}) and outline the training strategy (Sec. \ref{sec:method:train}).

\subsection{Content Anchors}
\label{sec:method:anchor}
We design three types of content anchors with different roles in supporting character video generation.
The global anchor $\mathcal{C}_g$ captures the overall scene, providing global subject–background correlation, it maintains the same resolution as the video sequence. 
The viewpoint anchor $\mathcal{C}_v$ represents multiple views of the character (up to $4$ in total), while the expression anchor $\mathcal{C}_e$ denotes diverse facial expressions (up to $8$ categories). These anchors constitute a structured and compact set to provide complementary cues for maintaining the overall scene, appearance across viewpoints, and expressive identities throughout the video.

\textbf{Superset Content Anchoring.}
During training, for each $5$s clip, we give arbitrary content anchors to provide corresponding references.
However, if all anchors are sampled only within the $5$s training clip, the model tends to exploit a shortcut: it simply reproduces these anchor frames in the generated sequence, leading to a copy–paste pattern rather than learning correlations between anchors and targets.
To prevent this issue, we extend the anchor selection range to the full-length source video. Specifically, for each clip, we trace it back to its original long-shot video and sample additional anchor frames outside the clip, as illustrated in Fig.~\ref{fig:method}.
This superset design introduces anchors with extra viewpoints, expressions, or poses for the training clip, encouraging the model to build correspondences among the generated sequence and anchor frames instead of solely copying.

To inject these anchors into the model, we first encode both the noised video sequence $x_t$ and all content anchors $\mathcal{C}_{g,v,e}$ through the same 3D VAE to obtain their latent tokens. 
Since the resolution of anchors and training clips may differ, each latent sequence is then patchified separately by a shared patchify layer $f_p(\cdot)$, and then concatenated before feeding them into the DiT backbone for self-attention computation. 
The first-frame latent $\mathcal{I}$ is placed at the beginning of the sequence as the global initialization, while the noised video latent $x_t$ represents the target to be denoised.

\textbf{RoPE as weak condition.}
After obtaining the concatenated latent sequence, we further introduce a RoPE-based mechanism to differentiate distinct anchors during self-attention computation.
Multiple content anchors serve distinct purposes but may share visual overlap, which causes attention conflicts when the model refers to them simultaneously \cite{ju2025fulldit,deng2025emerging}.
To disentangle their roles, we modify the 3D RoPE by assigning distinct temporal offsets to each anchor type, formulated as:
\begin{equation}
RoPE_i = RoPE(t + o_i + so_j, h_i, w_i),
\end{equation}
where $i \in \{l,v,e\}, j \in \{v,e\}$.
Here, $t$ corresponds to the temporal position of the generated video latent, while $o_i$ and $so_j$ denote the base and sub-offsets assigned to anchor types and their subcategories. 
We empirically set $o_v, o_e, o_l = 200, 400, 600$, respectively, 
and anchors of the same purpose are assigned sub-offsets $so_j\in[0,1,2,...,7]$ for different viewpoints (front, back, left, right) or expressions with varying semantics.
By binding each anchor to a unique RoPE offset, the model learns to treat 3D RoPE as a weak condition that implicitly encodes anchor identity, resolving content conflicts during self-attention.

\subsection{Multi-Modal Condition Injection}
\label{sec:method:audio}
The content-anchor mechanism is a general solution that can be seamlessly integrated with multi-modal conditions as control, such as text and audio. 
In this subsection, we describe the design of the audio branch, which enables synchronized and expressive character motion.

We employ wav2vec~\cite{baevski2020wav2vec} and a learnable projection layer to extract audio features $\mathbf{F}_{a} \in \mathbb{R}^{f \times d}$, where $f$ denotes the number of frames and $d$ the feature dimension. 
The audio features are injected via cross-attention layers, parallel to the text and image cross-attention, but equipped with a distinct projection-out layer to avoid feature interference. To ensure audio-video synchronization, we adopt local window attention for strict temporal alignment between the $\mathbf{F}_{a}$ and the video latent sequence. In detail, the $\mathbf{F}_{a}$ is reshaped into local windows, formulated as $\mathbf{F}^{'}_{a} \in \mathbb{R}^{N \times 4 \times d}$, where $N$ is the number of windows and $4$ indicates the frames per window, matching the temporal compression rate.

Finally, the video diffusion process follows flow matching~\cite{lipman2022flow}. Given a clean video $x_1$ and random noise $x_0 \sim \mathcal{N}(0, I)$, an intermediate state $x_t$ is interpolated as $x_t = t x_1 + (1 - t) x_0$, and the ground-truth velocity is computed as $v_t = \frac{d x_t}{dt} = x_1 - x_0$. The model predicts the velocity and is optimized through the following objective:
\begin{equation}
\label{eq:loss}
L = \mathbb{E}_{x_0, x_1, c, t} \left\| u(x_t, c, t; \theta) - v_t \right\|^2,
\end{equation}
where $c$ represents all conditional inputs, including image, text, audio, and content anchors, $\theta$ denotes the model parameters, and $u(x_t, c, t; \theta)$ is the predicted velocity.

\subsection{Inference}
\label{sec:method:infer}
During inference, the audio, $\mathcal{C}_v$, and $\mathcal{C}_e$ are optional inputs, and the number of anchor frames for $\mathcal{C}_v$ and $\mathcal{C}_e$ is flexible.  
When the $\mathcal{C}_v$ and $\mathcal{C}_e$ are provided, they support multi-view appearance and expressive identity consistency, while the input frame is treated as the global anchor $\mathcal{C}_g$.
To enable long-sequence generation, we adopt a chunk-wise autoregressive inference strategy.  
The video is divided into overlapping chunks.  
For each new chunk, the last four causal video latent from the previous chunk are used as prefix tokens to initialize the denoising process, ensuring temporal continuity between adjacent segments. 
A linear blending weight is applied to the overlapped regions, allowing smooth transitions between chunks.
All anchors remain fixedly attached throughout all chunks to serve as persistent references for consistency.

\subsection{Training Recipes}
\label{sec:method:train}
The model is trained in three stages, including the audio branch injection, introducing the global anchor, and injecting the viewpoint and expression anchors.

\textbf{Stage I}. 
We begin by establishing audio–visual alignment to enable synchronized and emotion-aware character generation.   
The model is trained on $2$M clips with text, image, and audio as inputs.  
To prevent model collapse at the initial stage, we initialize the value-projection weights of the audio branch to zero.  
During training, classifier-free guidance (CFG) \cite{ho2022classifier} is applied to both text and audio with a probability of $0.1$.  
After this stage, the model achieves precise lip synchronization, and common artifacts such as abrupt shot changes, subtitle overlays, and color shifts are largely removed.  

\textbf{Stage II.} 
In this stage, we further introduce the global anchor $\mathcal{C}_g$ to strengthen temporal coherence across extended sequences.  
To support the chunk-wise autoregressive inference described in Sec.~\ref{sec:method:infer}, we simulate the discontinuity between chunks during training. Specifically, With a certain probability, four ground-truth video latent are provided as prefix tokens at the beginning of each clip, mimicking the causal initialization during inference.  
This probabilistic prefix sampling enables the model to handle boundary transitions and maintain smooth continuity across chunks. 
The entire stage is trained on further $10$M clips. 

\textbf{Stage III.} 
Finally, we inject the viewpoint $\mathcal{C}_v$ and expression $\mathcal{C}_e$ anchors to encourage multi-view and expressive consistency.  
Because the data containing both types of anchors is rare, we first perform mixed training with samples that include only one anchor type, followed by fine-tuning on a smaller subset containing both.  
Statistical details for each anchor type and data ratio are provided in the Sup. 
This strategy, trained on $500$K clips, allows the model to extract relevant contents from diverse anchors and maintain consistent multi-view appearance and expression identity.  
\section{Experiment}
\textbf{Implementation.} The final model comprises $17$B parameters and includes $40$ DiT blocks, half of the blocks contain audio branch. We perform full-parameters training and the optimization employs the Adam \cite{kingma2014adam} with a constant learning rate of $1e-5$, $\beta=(0.9,0.95)$, and the weight decay is set to $0.1$. The entire training is conducted on $512$ A$800$ GPUs, with a total of $20,000$ training steps. During the inference, the denoising steps is set to $30$ with CFG $= 3.5$.

\begin{figure*}[t]
\centering
\begin{minipage}[c]{0.28\linewidth}
\centering
\small
\renewcommand{\arraystretch}{1.}
\renewcommand{\tabcolsep}{2 pt}
\captionof{table}{Quantitative comparison of long-term consistency.}
\begin{tabular}{c|ccc}
\toprule
\textbf{Method}         & \multicolumn{1}{c}{\textbf{Sub.}} $\uparrow$ & \textbf{Back.} $\uparrow$         & \textbf{Arcf.} $\uparrow$     \\ \midrule
HunyuanA.        &          0.914                       &               0.909        &         0.495                 \\
WanS2V         &          0.872                       &               0.864        &         0.439              \\
InfiniteTalk   &          0.952                       &               0.942        &         0.695                   \\
\rowcolor{mygray} \ Ours           &          \textbf{0.960}              &               \textbf{0.951}       &           \textbf{0.787}               \\ \bottomrule
\end{tabular}
\label{table:anchor_compare_long}
\end{minipage}
\hfill
\begin{minipage}[c]{0.31\linewidth}
\centering
\small
\renewcommand{\arraystretch}{1.}
\renewcommand{\tabcolsep}{1.8 pt}
\captionof{table}{Quantitative comparison of appearance and expressive ID consistency.}
\begin{tabular}{c|ccc}
\toprule
\textbf{Method}         & \multicolumn{1}{c}{\textbf{DINO-I}} $\uparrow$ & \textbf{CLIP-I} $\uparrow$ & \textbf{Exp.} $\uparrow$ \\ \midrule
Phantom        &      0.597                        &             0.785           &      0.645      \\
HunyuanC.     &      0.658                        &             0.767           &      0.511      \\
Vidu           &      0.662                        &             0.822           &      0.644      \\
Kling          &      0.698                       &             0.797           &      0.599      \\ 
Humo           &      0.810                        &             0.816           &      0.672      \\
\rowcolor{mygray} Ours          &      \textbf{0.821}                        &             \textbf{0.858}           &        \textbf{0.717}    \\ \bottomrule
\end{tabular}
\label{table:anchor_compare_view}
\end{minipage}
\hfill
\begin{minipage}[c]{0.4\linewidth}
\centering
\includegraphics[width=\linewidth]{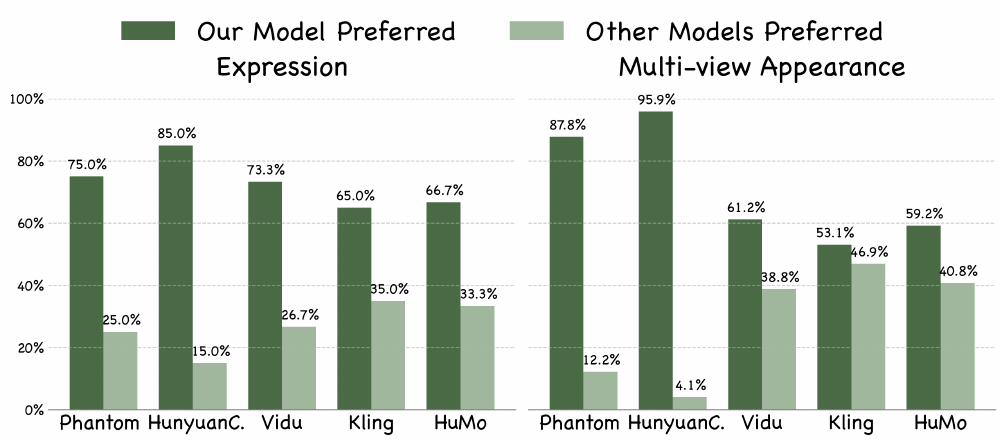}
\caption{The user study results of expressive ID and multi-view appearance consistency.}
\label{fig:userstudy_anchor}
\end{minipage}
\end{figure*}

\begin{figure*}[t]
    \centering
    \includegraphics[width=0.95\linewidth]{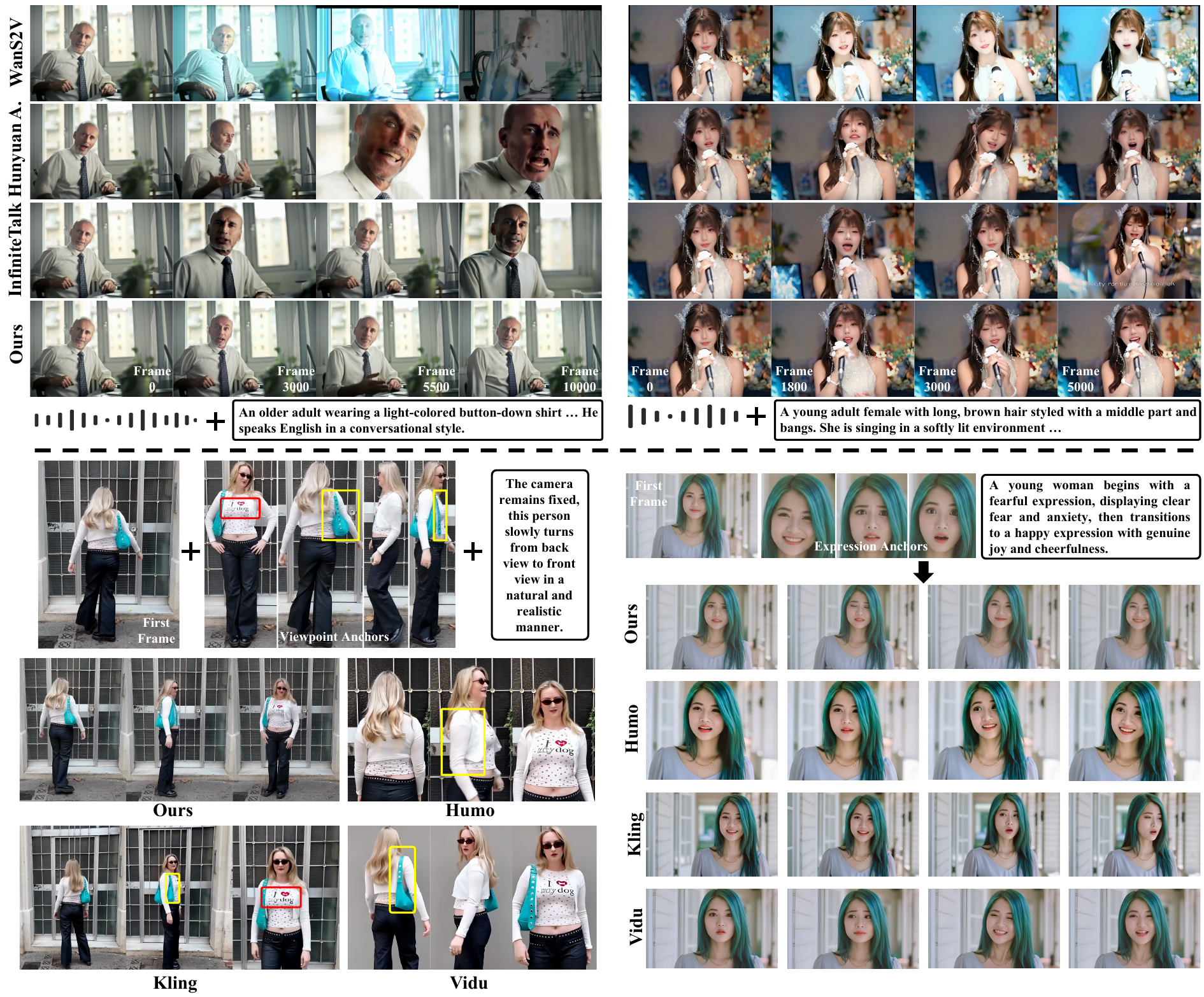}
    \caption{Qualitative comparison of long-term, multi-view appearance, and expressive identity consistency. The results show the top-3 performing methods and our method. The red and yellow boxes highlight appearance details.}
    \label{fig:compare}
\end{figure*}

\begin{figure*}[t]
\centering
\begin{minipage}[c]{0.45\linewidth}
\centering
\small
\captionof{table}{Quantitative comparison of fundamental capability.}
\renewcommand{\arraystretch}{1.}
\renewcommand{\tabcolsep}{3.5 pt}
\begin{tabular}{c|cccc}
\toprule
\textbf{Method}         & \multicolumn{1}{c}{\textbf{IQA}} $\uparrow$ & \textbf{AES} $\uparrow$        & \textbf{Sync-C} $\uparrow$      & \textbf{Sync-D} $\downarrow$ \\ \midrule
FantasyTalking        &           4.01         &         3.08        &              2.15       &  11.19 \\
HunyuanAvatar        &          4.41                       &               3.44        &          4.73                & 9.07  \\
WanS2V         &      4.49                             &                3.51      &              3.79             &  9.74 \\
InfiniteTalk   &           4.47                        &              3.50        &         4.48                  &  9.02 \\
OmniHuman1.5      &      4.58                             &            3.56          &          4.42                 & 9.52 \\
SekoTalk       &       4.62                            &             \textbf{3.65}         &        4.32                   & 9.14 \\
\rowcolor{mygray} Ours           &      \textbf{4.65}                             &               3.63       &           \textbf{5.12}                & \textbf{8.83} \\ \bottomrule
\end{tabular}
\label{table:base_compare}
\end{minipage}
\hfill
\begin{minipage}[c]{0.52\linewidth}
\centering
\includegraphics[width=\linewidth]{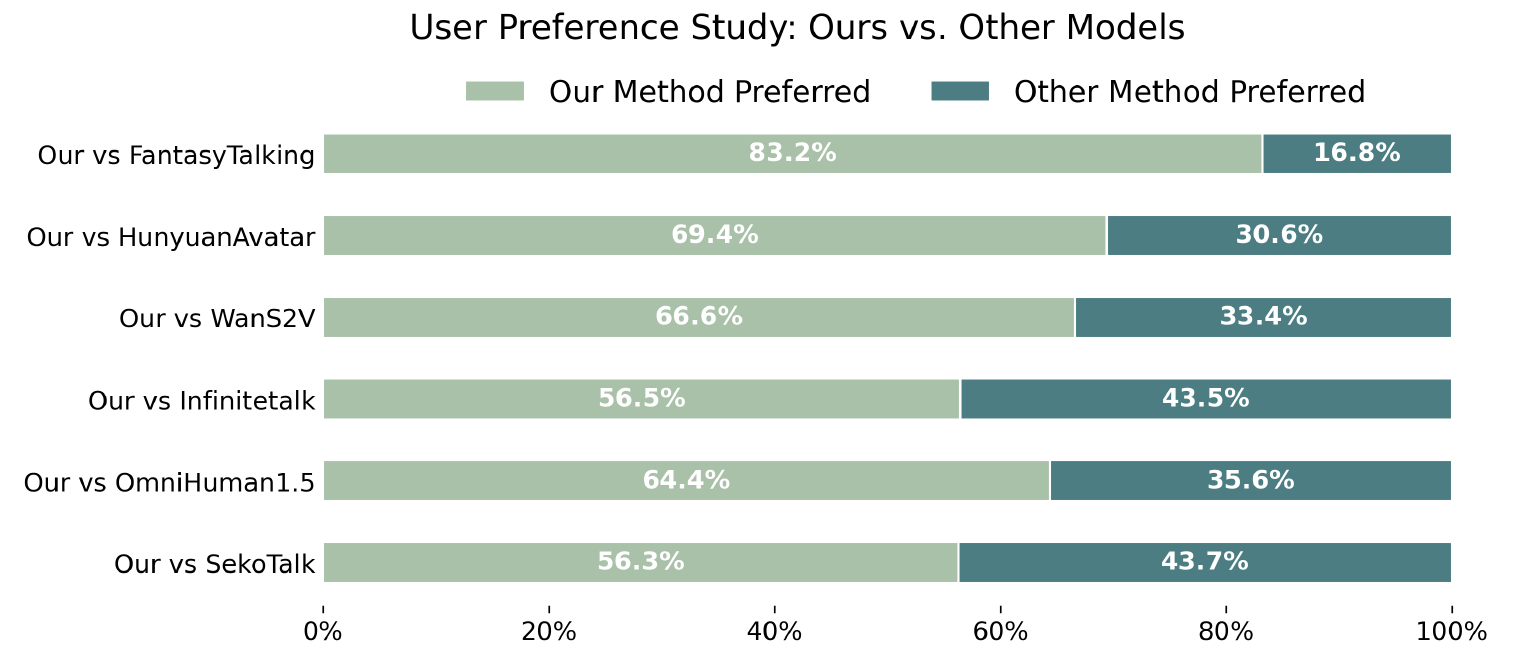}
\caption{The user study results of fundamental capability.}
\label{fig:userstudy}
\end{minipage}
\end{figure*}

\subsection{Benchmark}
\textbf{Baselines.} We employ DiT-based human-centric models that support long-duration inference for comparison in long-term consistency, including the InfiniteTalk \cite{yang2025infinitetalkaudiodrivenvideogeneration}, HunyuanAvatar \cite{hu2025HunyuanVideo-Avatar} (HunyuanA.), WanS2V \cite{gao2025wans2vaudiodrivencinematicvideo}. For the multi-view appearance, expressive facial performance, and ID preservation, we compare with multi-reference-based methods, including Humo \cite{25_humo}, Phantom \cite{liu2025phantom}, HunyuanCustom \cite{hu2025hunyuancustom} (HunyuanC.), Vidu-Q2 \cite{bao2024vidu} and Kling \cite{kuaishou2024kling}. In addition, we compare with several other human-centric models for fundamental capability evaluation, including FantasyTalking \cite{wang2025fantasytalking}, Omnihuman1.5 \cite{jiang2025omnihuman}, and SekoTalk \cite{SekoTalk}.

\textbf{Testset and Evaluation Metrics.} For long-term consistency evaluation, we make $20$ clips of various scenes, each lasting $5–10$ minutes, with an image, text and audio as inputs.
Regarding multi-view appearance consistency, we prepare $50$ characters with multi-view images and paired text prompts to show different views of the character.
For identity consistency across diverse expressions, we construct $40$ character sets, each containing $2–4$ expression keyframes with corresponding audio and prompt to drive the emotion change.
To evaluate the foundation ability, we gather $95$ cases demonstrated by various human-centric baselines, including the input image and audio, and we use the Gemini \cite{comanici2025gemini} to make the text prompt. We also test them on talking head dataset \eg, Celebv-HQ \cite{zhu2022celebvhq}, reported in the Sup.
Regarding the metrics, the subject and background consistency in VBench \cite{huang2023vbench} and the Arcface score \cite{deng2019arcface} between the first frame and the generated sequence are used to evaluate the long-term consistency. 
For the multi-view appearance and expressive identity, we use CLIP-I \cite{radford2021learning}, DINO-I \cite{simeoni2025dinov3}, and the Arcface score between the expression anchors and generated sequence (Exp.) to evaluate the consistency. In addition, the IQA, AES \cite{wu2023q} and Sync-C, Sync-D \cite{chung2016out} are used to evaluate the foundation ability.

\textbf{User Study.} We conduct the user study with $60$ domain experts from fields including film and digital arts, employing a blind, pairwise $A/B$ testing methodology. In each test, experts are presented with a pair of videos, one from our model and one from another method, with the random order and conceal model identities. For reference-based comparison, they select the superior video based on $3$ key dimensions, including video-reference correlation, natural degree, identity and appearance consistency, while for foundation comparison, they consider visual quality, motion quality, audio-video synchronization, please refer to Sup. for evaluation details. The final score 1 indicates a majority win, while 0 means defeat, and 0.5 is tie. We compute the final scores to represent overall human preference.

\begin{table}[t]
\centering
\small
  \renewcommand{\arraystretch}{1.}
  \renewcommand{\tabcolsep}{2 pt}
  \caption{Quantitative ablation results of content anchors ($\mathcal{C}_{g,v,e}$) and the Superset Content Anchoring (SCA) and Rope as Weak Condition (RWC) mechanisms.}
\label{table:ablation}
\begin{tabular}{cccc|cccc}
\toprule
\multicolumn{4}{c|}{\textbf{Long-term}}                            & \multicolumn{4}{c}{\textbf{Appearance $\&$ ID}}                             \\ \midrule
\multicolumn{1}{c|}{\textbf{Setting}}       & \textbf{Sub.} & \textbf{Back.} & \textbf{Arcf.} & \multicolumn{1}{c|}{\textbf{Setting}}  & \textbf{DINO-I} & \textbf{CLIP-I} & \textbf{Exp.}\\ \midrule
\multicolumn{1}{c|}{Ours}      &             \textbf{0.960}         &        \textbf{0.951}     &    \textbf{0.787}  &         \multicolumn{1}{c|}{Ours}      &             \textbf{0.821}         &        \textbf{0.858}     &    \textbf{0.717}    \\
\multicolumn{1}{c|}{\ding{55} $\mathcal{C}_g$}      &       0.840        &    0.844      &    0.233  & \multicolumn{1}{c|}{\ding{55} $\mathcal{C}_{v,e}$}  &       0.801        &    0.848      &    0.668\\
\multicolumn{1}{c|}{\ding{55} SCA}       &            0.922          &       0.899       &    0.489    & \multicolumn{1}{c|}{\ding{55} RWC}       &          0.817       &     0.855          & 0.671\\ \bottomrule
\end{tabular}
\end{table}

\subsection{Comparison}
The quantitative comparison results of long-term consistency are reported in Tab. \ref{table:anchor_compare_long}, the multi-view appearance and expressive identity consistency are presented in Tab. \ref{table:anchor_compare_view}. Our method outperforms other methods across metrics. Meanwhile, the user study (Fig. \ref{fig:userstudy_anchor}) further demonstrates that our method is consistently preferred over existing approaches.
We also provide some qualitative comparisons in Fig. \ref{fig:compare}. Regarding the long-term consistency, other methods exhibit obvious drifting across both the overall scene and character identity. The InfiniteTalk maintains overall scene by repeatedly inserting the first frame into the generated sequence, but its character identity tends to degrade over the chunk (reflected in a low score of Arcface). Moreover, this manner introduces noticeable frame repetitions, also shown in the figure. For multi-view appearance and facial expressions, our method better preserves fine-grained details and effectively drives the character to perform corresponding contents. The quantitative comparison results for the foundation capability are presented in Tab. \ref{table:base_compare}, and the corresponding user study results are shown in Fig. \ref{fig:userstudy}.

\subsection{Ablations}
\begin{figure}
    \centering
    \begin{overpic}[width=1.0\linewidth]{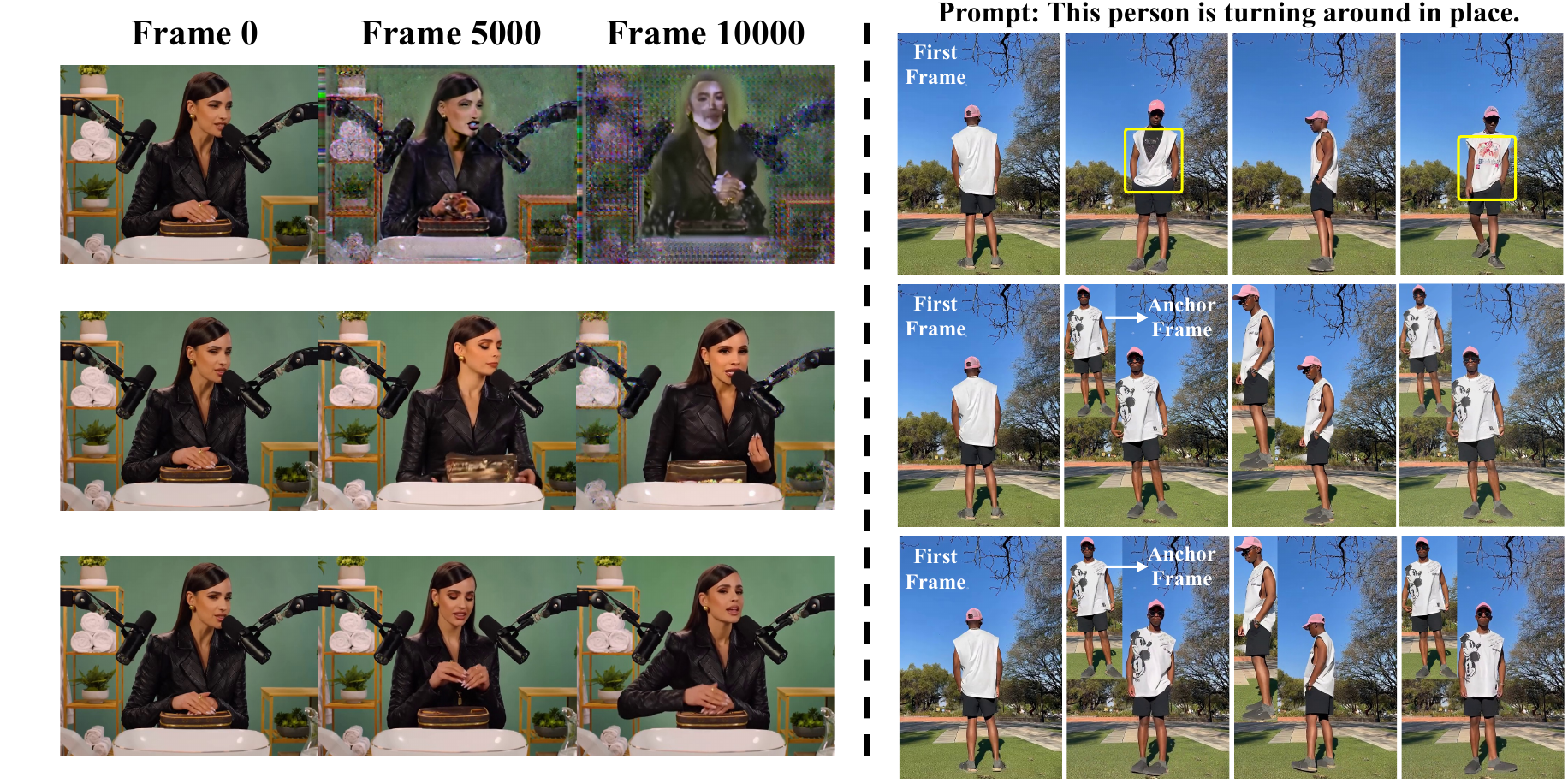}
    \put(28,-1.3){{\textbf{\footnotesize (a)}}}
    \put(76.5,-2.5){{\textbf{\footnotesize (b)}}}

    \put(0,33.5){\rotatebox{90}{\textbf{\scriptsize \ding{55} anchor}}}
    \put(0,17){\rotatebox{90}{\textbf{\scriptsize $w/o$ SCA}}}
    \put(0,4){\rotatebox{90}{\textbf{\scriptsize Ours}}}
    
    \end{overpic}
    \caption{\textbf{(a)} Ablation of the global anchor ($\mathcal{C}_g$). \textbf{(b)} Ablation of the ($\mathcal{C}_v$). The top row indicates $w/o$ the $\mathcal{C}_v$, the middle row indicates $w$ the $\mathcal{C}_v$ but $w/o$ the Superset Content Anchoring (SCA), and the bottom row indicates with both. The small images at the top-left corner of each frame indicate the input $\mathcal{C}_v$.}
    \label{fig:ablation1}
\end{figure}

The quantitative ablation results on global, viewpoint, and expression anchors, as well as the Superset Content Anchoring (SCA) and RoPE as Weak Condition (RWC) strategy, are reported in Tab. \ref{table:ablation}. Further, we provide qualitative results to offer a more intuitive comparison. As shown in Fig. \ref{fig:ablation1}, when there is no global anchor, the long video exhibits obvious drift. Similarly, without viewpoint anchors, the character displays inconsistent appearance \eg, the logo on the clothing. When anchors are provided but the SCA strategy is not applied, long-term consistency improves noticeably, but slight color deviations remain, and the model tends to directly copy the viewpoint anchor frames, including the human pose. In contrast, combining both content anchors and the SCA strategy enables the model to maintain strong long-term consistency and extract relevant appearance from the anchors rather than directly copying them.

\begin{figure}
    \centering
    \includegraphics[width=0.95\linewidth]{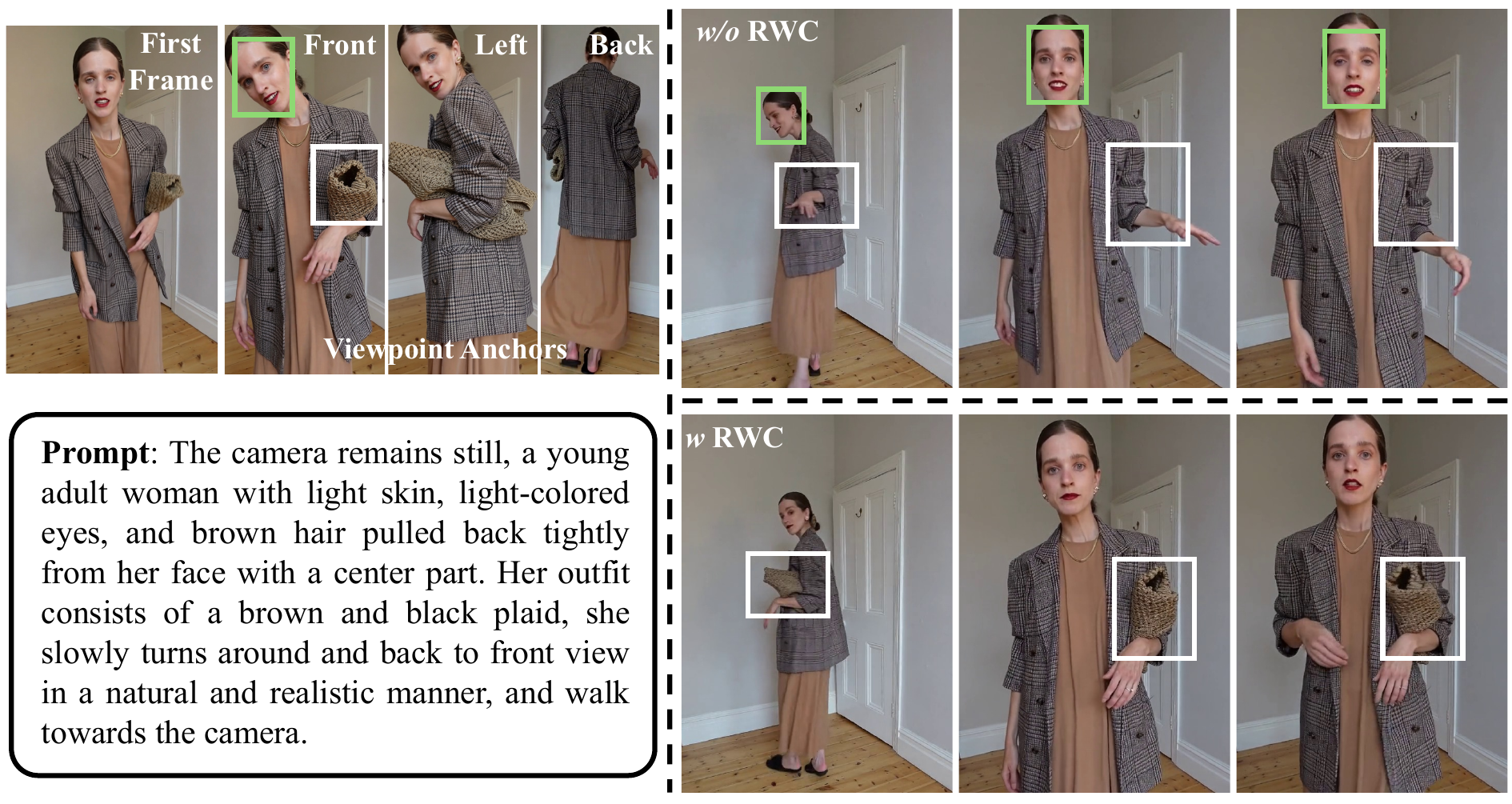}
    \caption{The left includes the first frame, viewpoint anchor frames and the text prompt. The right side shows generated videos, the bottom row indicates using ``RoPE as Weak Condition'' (RWC) while the top row dose not employ it.}
    \label{fig:ablation2}
\end{figure}

Qualitative results of ``RoPE as weak condition'' are shown in Fig. \ref{fig:ablation2}. When the RoPE offset is not set to segment different anchors, the reference exhibits conflicts. As can be seen, when viewpoint anchors are provided but contains certain facial expressions, the model partially refers the expression \eg, smile, while neglecting some appearance details, such as the object held in the hand. In contrast, when the RoPE offset is set as a weak condition, the model distinguishes anchors with different purposes and extracts the corresponding content from them.

\subsection{Analysis}
\begin{figure}
    \centering
    \includegraphics[width=0.9\linewidth]{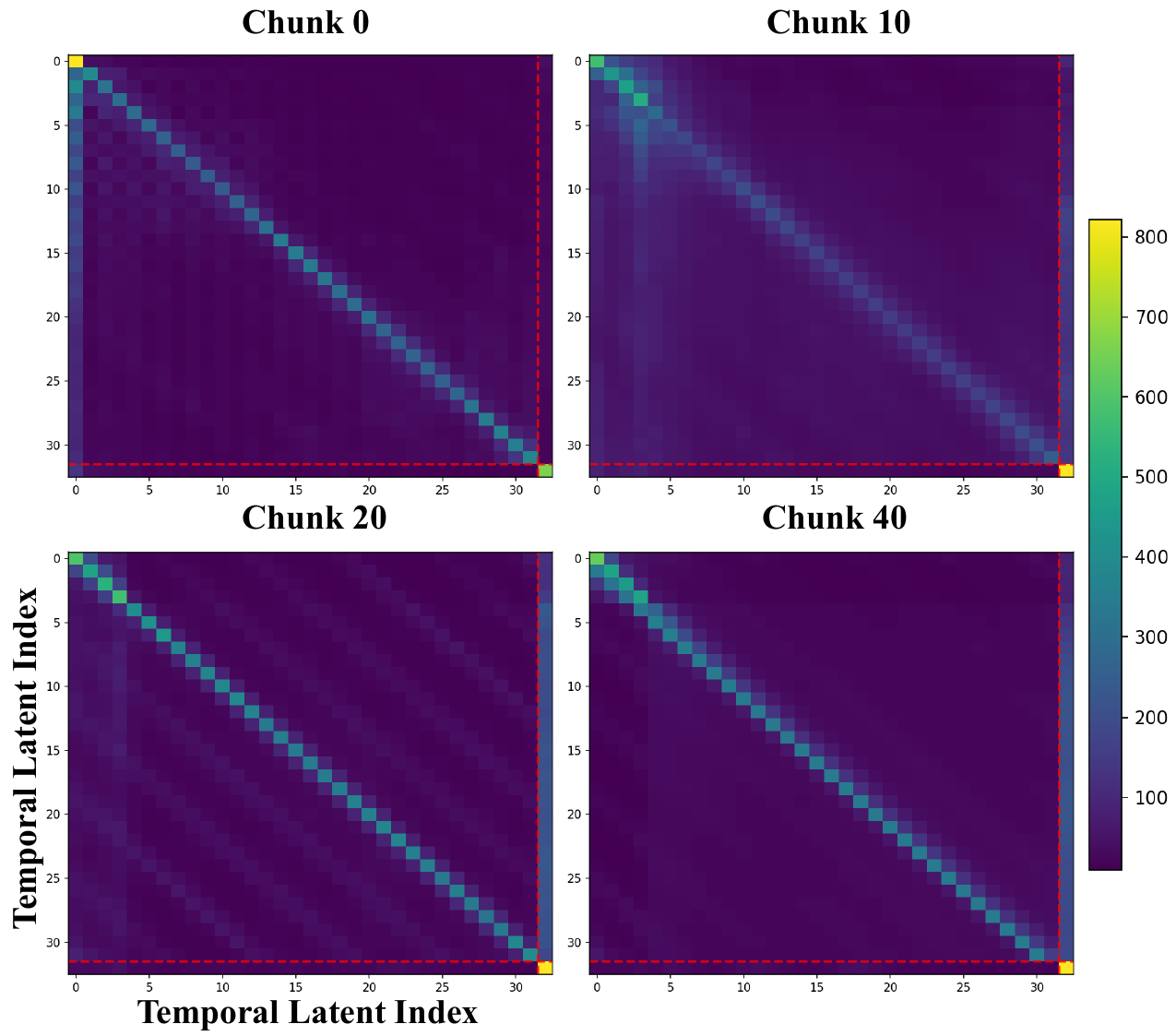}
    \caption{Self-attention maps of the generated sequence and its attention toward the global anchor across different chunks (each lasting $5$s). The rightmost dashed column indicates the attention score from the generated sequence to the global anchor frame.}
    \label{fig:ana_atten_map}
\end{figure}

We further analyze the intermediate representations within the model to validate the effectiveness of the ``content anchor'' design. For long-term consistency, Fig. \ref{fig:ana_atten_map} visualizes the attention maps of the generated sequence across different chunks during long-sequence inference, including its self-attention and the attention toward the global anchor. As the number of chunks increases, the attention to the beginning latent gradually decreases, meanwhile, the attention toward the global anchor progressively strengthens, which indicates that the global anchor frame provides a stable reference that helps the model maintain long-term consistency.

\begin{figure}
    \centering
    \includegraphics[width=0.95\linewidth]{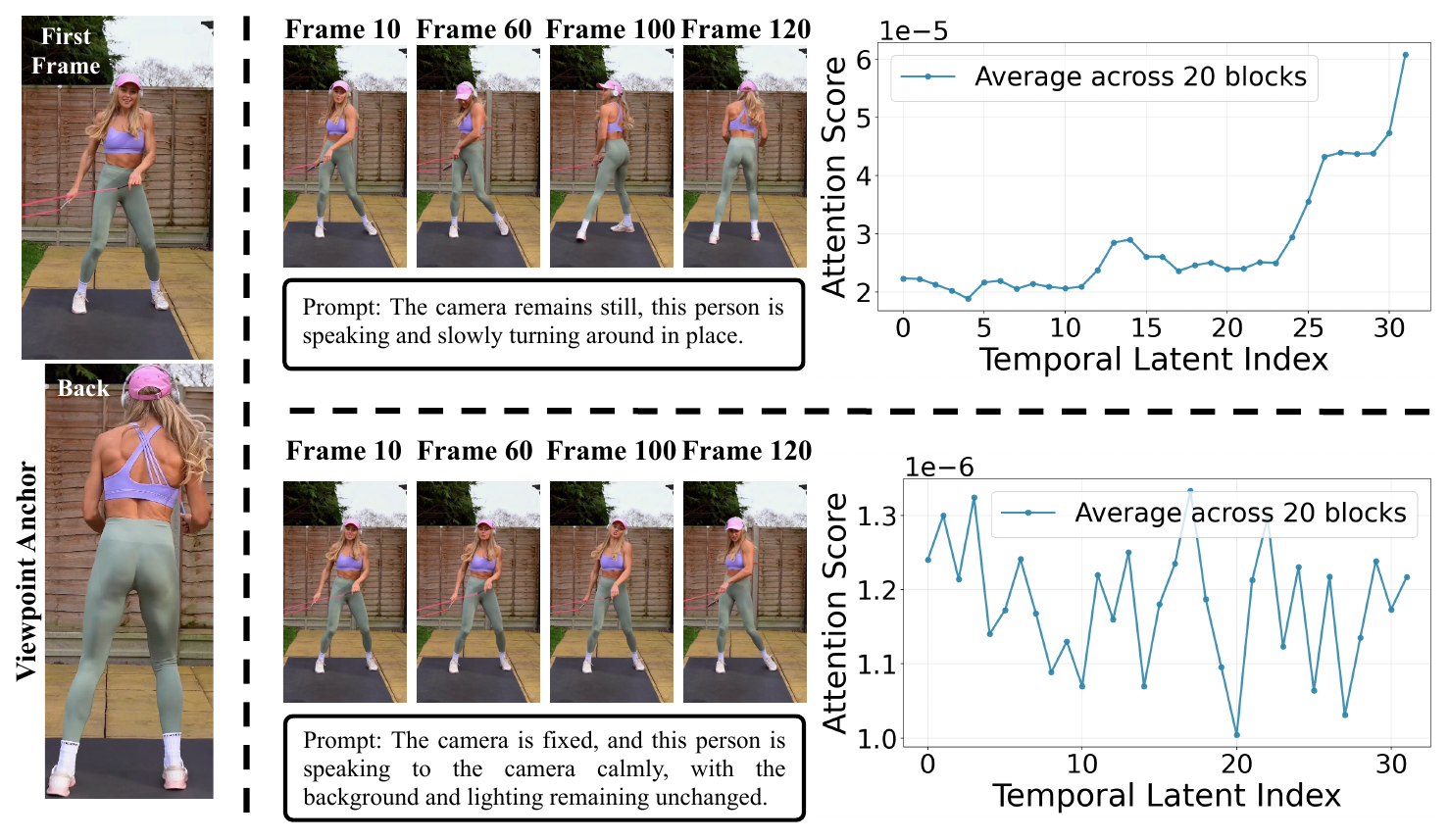}
    \caption{\textbf{Left:} the first frame and a back-view anchor frame. \textbf{Right:} the generated results and the corresponding attention score curve of the generated sequence toward the back-view anchor. The upper and lower rows show results generated under different texts.}

    \label{fig:ana_atten_curve}
\end{figure}

Moreover, as illustrated in Fig. \ref{fig:ana_atten_curve}, we use a front-view as the input and a back-view as the viewpoint anchor, while controlling the character with different text prompts, such as ``turning around" or ``speaking to the camera". The generated results indeed follow these instructions, and we visualize the attention score of the generated sequence toward the anchor frame to show the underlying mechanism.
As can be seen, the attention score significantly increases when the character turns around to face backward, while it remains steady and low score when the character is not instructed to rotate. This indicates that the model builds the correlation between the generated sequence and the content anchor, it adaptively extracts relevant information.

\section{Conclusion}
In this work, we propose Gloria, which leverages content anchors to generate long-duration character videos with consistent appearance and expressive identity. To address the key technical challenges in an anchor-based manner, we further introduce ``Superset Content Anchoring'' and ``RoPE as Weak Condition'', which mitigate copy-paste patterns and resolve conflicts among multiple anchor frames.
Moreover, to build the content anchor set, we develop a scalable data pipeline that automatically extracts anchor frames from large-scale video datasets.
Eventually, Gloria produces high-quality character videos exceeding $10$ minutes, maintaining strong consistency in identity and appearance.

\noindent\textbf{Acknowledgement} This work is supported by the National Natural Science Foundation of China (NSFC) under Grants 625B2175, 62225207, 62436008, 62306295, and 62576328.

{
    \small
    \bibliographystyle{ieeenat_fullname}
    \bibliography{main}

\begin{thebibliography}{96}
\providecommand{\natexlab}[1]{#1}
\providecommand{\url}[1]{\texttt{#1}}
\expandafter\ifx\csname urlstyle\endcsname\relax
  \providecommand{\doi}[1]{doi: #1}\else
  \providecommand{\doi}{doi: \begingroup \urlstyle{rm}\Url}\fi

\bibitem[Sek()]{SekoTalk}
Sekotalk.
\newblock \url{https://sekotalk.com/}.
\newblock Accessed: 2025-11-12.

\bibitem[Baevski et~al.(2020)Baevski, Zhou, Mohamed, and Auli]{baevski2020wav2vec}
Alexei Baevski, Yuhao Zhou, Abdelrahman Mohamed, and Michael Auli.
\newblock wav2vec 2.0: A framework for self-supervised learning of speech representations.
\newblock \emph{Advances in neural information processing systems}, 33:\penalty0 12449--12460, 2020.

\bibitem[Bao et~al.(2024)Bao, Xiang, Yue, He, Zhu, Zheng, Zhao, Liu, Wang, and Zhu]{bao2024vidu}
Fan Bao, Chendong Xiang, Gang Yue, Guande He, Hongzhou Zhu, Kaiwen Zheng, Min Zhao, Shilong Liu, Yaole Wang, and Jun Zhu.
\newblock Vidu: a highly consistent, dynamic and skilled text-to-video generator with diffusion models.
\newblock \emph{arXiv preprint arXiv:2405.04233}, 2024.

\bibitem[Brooks et~al.(2024)Brooks, Peebles, Holmes, DePue, Guo, Jing, Schnurr, Taylor, Luhman, Luhman, et~al.]{brooks2024video}
Tim Brooks, Bill Peebles, Connor Holmes, Will DePue, Yufei Guo, Li Jing, David Schnurr, Joe Taylor, Troy Luhman, Eric Luhman, et~al.
\newblock Video generation models as world simulators.
\newblock \emph{OpenAI Blog}, 1\penalty0 (8):\penalty0 1, 2024.

\bibitem[Chen et~al.(2025{\natexlab{a}})Chen, Chen, Xu, Li, Dong, Sun, Jiang, Li, Yang, Zhao, et~al.]{chen2025dancetogether}
Junhao Chen, Mingjin Chen, Jianjin Xu, Xiang Li, Junting Dong, Mingze Sun, Puhua Jiang, Hongxiang Li, Yuhang Yang, Hao Zhao, et~al.
\newblock Dancetogether! identity-preserving multi-person interactive video generation.
\newblock \emph{arXiv preprint arXiv:2505.18078}, 2025{\natexlab{a}}.

\bibitem[Chen et~al.(2025{\natexlab{b}})Chen, Ma, Liu, Li, Chen, Liu, He, Li, He, and Wu]{25_humo}
Liyang Chen, Tianxiang Ma, Jiawei Liu, Bingchuan Li, Zhuowei Chen, Lijie Liu, Xu He, Gen Li, Qian He, and Zhiyong Wu.
\newblock Humo: Human-centric video generation via collaborative multi-modal conditioning.
\newblock \emph{arXiv preprint arXiv:2509.08519}, 2025{\natexlab{b}}.

\bibitem[Chen et~al.(2025{\natexlab{c}})Chen, Siarohin, Menapace, Fang, Lee, Skorokhodov, Aberman, Zhu, Yang, and Tulyakov]{chen2025multi}
Tsai-Shien Chen, Aliaksandr Siarohin, Willi Menapace, Yuwei Fang, Kwot~Sin Lee, Ivan Skorokhodov, Kfir Aberman, Jun-Yan Zhu, Ming-Hsuan Yang, and Sergey Tulyakov.
\newblock Multi-subject open-set personalization in video generation.
\newblock In \emph{Proceedings of the Computer Vision and Pattern Recognition Conference}, pages 6099--6110, 2025{\natexlab{c}}.

\bibitem[Chen et~al.(2024)Chen, Liu, Chen, Feng, Liu, Shen, and Zhao]{chen2024livephoto}
Xi Chen, Zhiheng Liu, Mengting Chen, Yutong Feng, Yu Liu, Yujun Shen, and Hengshuang Zhao.
\newblock Livephoto: Real image animation with text-guided motion control.
\newblock In \emph{European Conference on Computer Vision}, pages 475--491. Springer, 2024.

\bibitem[Chen et~al.(2025{\natexlab{d}})Chen, Hu, Zhang, Xue, Yi, and Ma]{chen2025instancev}
Yuheng Chen, Teng Hu, Jiangning Zhang, Zhucun Xue, Ran Yi, and Lizhuang Ma.
\newblock Instancev: Instance-level video generation.
\newblock \emph{arXiv preprint arXiv:2511.23146}, 2025{\natexlab{d}}.

\bibitem[Chen et~al.(2025{\natexlab{e}})Chen, Liang, Zhou, Huang, Ma, Tang, Lin, Zhou, and Lu]{hu2025HunyuanVideo-Avatar}
Yi Chen, Sen Liang, Zixiang Zhou, Ziyao Huang, Yifeng Ma, Junshu Tang, Qin Lin, Yuan Zhou, and Qinglin Lu.
\newblock Hunyuanvideo-avatar: High-fidelity audio-driven human animation for multiple characters, 2025{\natexlab{e}}.

\bibitem[Chen et~al.(2025{\natexlab{f}})Chen, Cao, Chen, Li, and Ma]{25aaai_echomimic}
Zhiyuan Chen, Jiajiong Cao, Zhiquan Chen, Yuming Li, and Chenguang Ma.
\newblock Echomimic: Lifelike audio-driven portrait animations through editable landmark conditions.
\newblock In \emph{Proceedings of the AAAI Conference on Artificial Intelligence}, pages 2403--2410, 2025{\natexlab{f}}.

\bibitem[Chung and Zisserman(2016{\natexlab{a}})]{Chung16a}
J.~S. Chung and A. Zisserman.
\newblock Out of time: automated lip sync in the wild.
\newblock In \emph{Workshop on Multi-view Lip-reading, ACCV}, 2016{\natexlab{a}}.

\bibitem[Chung and Zisserman(2016{\natexlab{b}})]{chung2016out}
Joon~Son Chung and Andrew Zisserman.
\newblock Out of time: automated lip sync in the wild.
\newblock In \emph{Asian conference on computer vision}, pages 251--263. Springer, 2016{\natexlab{b}}.

\bibitem[Comanici et~al.(2025)Comanici, Bieber, Schaekermann, Pasupat, Sachdeva, Dhillon, Blistein, Ram, Zhang, Rosen, et~al.]{comanici2025gemini}
Gheorghe Comanici, Eric Bieber, Mike Schaekermann, Ice Pasupat, Noveen Sachdeva, Inderjit Dhillon, Marcel Blistein, Ori Ram, Dan Zhang, Evan Rosen, et~al.
\newblock Gemini 2.5: Pushing the frontier with advanced reasoning, multimodality, long context, and next generation agentic capabilities.
\newblock \emph{arXiv preprint arXiv:2507.06261}, 2025.

\bibitem[Deng et~al.(2025)Deng, Zhu, Li, Gou, Li, Wang, Zhong, Yu, Nie, Song, et~al.]{deng2025emerging}
Chaorui Deng, Deyao Zhu, Kunchang Li, Chenhui Gou, Feng Li, Zeyu Wang, Shu Zhong, Weihao Yu, Xiaonan Nie, Ziang Song, et~al.
\newblock Emerging properties in unified multimodal pretraining.
\newblock \emph{arXiv preprint arXiv:2505.14683}, 2025.

\bibitem[Deng et~al.(2019)Deng, Guo, Xue, and Zafeiriou]{deng2019arcface}
Jiankang Deng, Jia Guo, Niannan Xue, and Stefanos Zafeiriou.
\newblock Arcface: Additive angular margin loss for deep face recognition.
\newblock In \emph{Proceedings of the IEEE/CVF conference on computer vision and pattern recognition}, pages 4690--4699, 2019.

\bibitem[Ding et~al.(2025)Ding, Liu, Zhang, Wang, Hu, Cui, Lao, Shao, Liu, Li, et~al.]{ding2025kling}
Yikang Ding, Jiwen Liu, Wenyuan Zhang, Zekun Wang, Wentao Hu, Liyuan Cui, Mingming Lao, Yingchao Shao, Hui Liu, Xiaohan Li, et~al.
\newblock Kling-avatar: Grounding multimodal instructions for cascaded long-duration avatar animation synthesis.
\newblock \emph{arXiv preprint arXiv:2509.09595}, 2025.

\bibitem[Fan et~al.(2024{\natexlab{a}})Fan, Tang, Cao, Yi, Li, Gong, Zhang, Wang, Wang, and Ma]{fan2024freemotion}
Ke Fan, Junshu Tang, Weijian Cao, Ran Yi, Moran Li, Jingyu Gong, Jiangning Zhang, Yabiao Wang, Chengjie Wang, and Lizhuang Ma.
\newblock Freemotion: A unified framework for number-free text-to-motion synthesis.
\newblock In \emph{European Conference on Computer Vision}, pages 93--109. Springer Nature Switzerland Cham, 2024{\natexlab{a}}.

\bibitem[Fan et~al.(2024{\natexlab{b}})Fan, Zhang, Yi, Gong, Wang, Wang, Tan, Wang, and Ma]{fan2024textual}
Ke Fan, Jiangning Zhang, Ran Yi, Jingyu Gong, Yabiao Wang, Yating Wang, Xin Tan, Chengjie Wang, and Lizhuang Ma.
\newblock Textual decomposition then sub-motion-space scattering for open-vocabulary motion generation.
\newblock \emph{arXiv preprint arXiv:2411.04079}, 2024{\natexlab{b}}.

\bibitem[Fan et~al.(2025)Fan, Lu, Dai, Yu, Xiao, Dou, Dong, Ma, and Wang]{fan2025go}
Ke Fan, Shunlin Lu, Minyue Dai, Runyi Yu, Lixing Xiao, Zhiyang Dou, Junting Dong, Lizhuang Ma, and Jingbo Wang.
\newblock Go to zero: Towards zero-shot motion generation with million-scale data.
\newblock In \emph{Proceedings of the IEEE/CVF International Conference on Computer Vision}, pages 13336--13348, 2025.

\bibitem[Fei et~al.(2025)Fei, Jiang, Qiu, Gu, Zhang, Wang, Bai, Li, Fan, Chen, et~al.]{fei2025skyreels}
Zhengcong Fei, Hao Jiang, Di Qiu, Baoxuan Gu, Youqiang Zhang, Jiahua Wang, Jialin Bai, Debang Li, Mingyuan Fan, Guibin Chen, et~al.
\newblock Skyreels-audio: Omni audio-conditioned talking portraits in video diffusion transformers.
\newblock \emph{arXiv preprint arXiv:2506.00830}, 2025.

\bibitem[Gan et~al.(2025)Gan, Yang, Zhu, Xue, and Hoi]{gan2025omniavatar}
Qijun Gan, Ruizi Yang, Jianke Zhu, Shaofei Xue, and Steven Hoi.
\newblock Omniavatar: Efficient audio-driven avatar video generation with adaptive body animation, 2025.

\bibitem[Gao et~al.(2025)Gao, Hu, Hu, Huang, Ji, Meng, Qi, Qiao, Shen, Song, Sun, Tian, Wang, Wang, Wang, Xiao, Xu, Zhang, Zhang, Zhang, Zhang, Zhou, and Zhuo]{gao2025wans2vaudiodrivencinematicvideo}
Xin Gao, Li Hu, Siqi Hu, Mingyang Huang, Chaonan Ji, Dechao Meng, Jinwei Qi, Penchong Qiao, Zhen Shen, Yafei Song, Ke Sun, Linrui Tian, Guangyuan Wang, Qi Wang, Zhongjian Wang, Jiayu Xiao, Sheng Xu, Bang Zhang, Peng Zhang, Xindi Zhang, Zhe Zhang, Jingren Zhou, and Lian Zhuo.
\newblock Wan-s2v: Audio-driven cinematic video generation, 2025.

\bibitem[{Google AI Studio}(2025)]{google_gemini_flash_image}
{Google AI Studio}.
\newblock Gemini 2.5 flash image.
\newblock \url{https://aistudio.google.com/models/gemini-2-5-flash-image}, 2025.
\newblock Accessed: 2025-11-14.

\bibitem[HaCohen et~al.(2024{\natexlab{a}})HaCohen, Chiprut, Brazowski, Shalem, Moshe, Richardson, Levin, Shiran, Zabari, Gordon, Panet, Weissbuch, Kulikov, Bitterman, Melumian, and Bibi]{HaCohen2024LTXVideo}
Yoav HaCohen, Nisan Chiprut, Benny Brazowski, Daniel Shalem, Dudu Moshe, Eitan Richardson, Eran Levin, Guy Shiran, Nir Zabari, Ori Gordon, Poriya Panet, Sapir Weissbuch, Victor Kulikov, Yaki Bitterman, Zeev Melumian, and Ofir Bibi.
\newblock Ltx-video: Realtime video latent diffusion.
\newblock \emph{arXiv preprint arXiv:2501.00103}, 2024{\natexlab{a}}.

\bibitem[HaCohen et~al.(2024{\natexlab{b}})HaCohen, Chiprut, Brazowski, Shalem, Moshe, Richardson, Levin, Shiran, Zabari, Gordon, et~al.]{hacohen2024ltx}
Yoav HaCohen, Nisan Chiprut, Benny Brazowski, Daniel Shalem, Dudu Moshe, Eitan Richardson, Eran Levin, Guy Shiran, Nir Zabari, Ori Gordon, et~al.
\newblock Ltx-video: Realtime video latent diffusion.
\newblock \emph{arXiv preprint arXiv:2501.00103}, 2024{\natexlab{b}}.

\bibitem[Han et~al.(2025)Han, Zhai, Yang, Cao, and Zha]{han2025touch}
Guangyi Han, Wei Zhai, Yuhang Yang, Yang Cao, and Zheng-Jun Zha.
\newblock Touch: Text-guided controllable generation of free-form hand-object interactions.
\newblock \emph{arXiv preprint arXiv:2510.14874}, 2025.

\bibitem[Ho and Salimans(2022)]{ho2022classifier}
Jonathan Ho and Tim Salimans.
\newblock Classifier-free diffusion guidance.
\newblock \emph{arXiv preprint arXiv:2207.12598}, 2022.

\bibitem[Hogue et~al.(2024)Hogue, Zhang, Daruger, Tian, and Guo]{hogue2024diffted}
Steven Hogue, Chenxu Zhang, Hamza Daruger, Yapeng Tian, and Xiaohu Guo.
\newblock Diffted: One-shot audio-driven ted talk video generation with diffusion-based co-speech gestures.
\newblock In \emph{Proceedings of the IEEE/CVF Conference on Computer Vision and Pattern Recognition}, pages 1922--1931, 2024.

\bibitem[Hu et~al.(2023)Hu, Gao, Zhang, Sun, Zhang, and Bo]{23_animateanyone}
Li Hu, Xin Gao, Peng Zhang, Ke Sun, Bang Zhang, and Liefeng Bo.
\newblock Animate anyone: Consistent and controllable image-to-video synthesis for character animation.
\newblock \emph{arXiv preprint arXiv:2311.17117}, 2023.

\bibitem[Hu et~al.(2025{\natexlab{a}})Hu, Wang, Shen, Gao, Meng, Zhuo, Zhang, Zhang, and Bo]{25_animateanyone2}
Li Hu, Guangyuan Wang, Zhen Shen, Xin Gao, Dechao Meng, Lian Zhuo, Peng Zhang, Bang Zhang, and Liefeng Bo.
\newblock Animate anyone 2: High-fidelity character image animation with environment affordance.
\newblock \emph{arXiv preprint arXiv:2502.06145}, 2025{\natexlab{a}}.

\bibitem[Hu et~al.()Hu, Yu, Zhou, Liang, Lin, Zhou, Yi, and Lu]{huomnicustom}
Teng Hu, Zhentao Yu, Zhengguang Zhou, Sen Liang, Qin Lin, Yuan Zhou, Ran Yi, and Qinglin Lu.
\newblock Omnicustom: A multimodal-driven architecture for customized video generation.

\bibitem[Hu et~al.(2024)Hu, Zhang, Yi, Wang, Huang, Weng, Wang, and Ma]{hu2024motionmaster}
Teng Hu, Jiangning Zhang, Ran Yi, Yating Wang, Hongrui Huang, Jieyu Weng, Yabiao Wang, and Lizhuang Ma.
\newblock Motionmaster: Training-free camera motion transfer for video generation.
\newblock \emph{arXiv preprint arXiv:2404.15789}, 2024.

\bibitem[Hu et~al.(2025{\natexlab{b}})Hu, Yu, Zhou, Liang, Zhou, Lin, and Lu]{hu2025hunyuancustom}
Teng Hu, Zhentao Yu, Zhengguang Zhou, Sen Liang, Yuan Zhou, Qin Lin, and Qinglin Lu.
\newblock Hunyuancustom: A multimodal-driven architecture for customized video generation, 2025{\natexlab{b}}.

\bibitem[Hu et~al.(2025{\natexlab{c}})Hu, Yu, Zhou, Zhang, Zhou, Lu, and Yi]{hu2025polyvivid}
Teng Hu, Zhentao Yu, Zhengguang Zhou, Jiangning Zhang, Yuan Zhou, Qinglin Lu, and Ran Yi.
\newblock Polyvivid: Vivid multi-subject video generation with cross-modal interaction and enhancement.
\newblock \emph{arXiv preprint arXiv:2506.07848}, 2025{\natexlab{c}}.

\bibitem[Huang et~al.(2024)Huang, He, Yu, Zhang, Si, Jiang, Zhang, Wu, Jin, Chanpaisit, Wang, Chen, Wang, Lin, Qiao, and Liu]{huang2023vbench}
Ziqi Huang, Yinan He, Jiashuo Yu, Fan Zhang, Chenyang Si, Yuming Jiang, Yuanhan Zhang, Tianxing Wu, Qingyang Jin, Nattapol Chanpaisit, Yaohui Wang, Xinyuan Chen, Limin Wang, Dahua Lin, Yu Qiao, and Ziwei Liu.
\newblock {VBench}: Comprehensive benchmark suite for video generative models.
\newblock In \emph{Proceedings of the IEEE/CVF Conference on Computer Vision and Pattern Recognition}, 2024.

\bibitem[Hurst et~al.(2024)Hurst, Lerer, Goucher, Perelman, Ramesh, Clark, Ostrow, Welihinda, Hayes, Radford, et~al.]{gpt4o}
Aaron Hurst, Adam Lerer, Adam~P Goucher, Adam Perelman, Aditya Ramesh, Aidan Clark, AJ Ostrow, Akila Welihinda, Alan Hayes, Alec Radford, et~al.
\newblock Gpt-4o system card.
\newblock \emph{arXiv preprint arXiv:2410.21276}, 2024.

\bibitem[Jiang et~al.(2025{\natexlab{a}})Jiang, Zeng, Zheng, Yang, Liang, Liao, Liang, Zhang, and Gao]{jiang2025omnihuman}
Jianwen Jiang, Weihong Zeng, Zerong Zheng, Jiaqi Yang, Chao Liang, Wang Liao, Han Liang, Yuan Zhang, and Mingyuan Gao.
\newblock Omnihuman-1.5: Instilling an active mind in avatars via cognitive simulation.
\newblock \emph{arXiv preprint arXiv:2508.19209}, 2025{\natexlab{a}}.

\bibitem[Jiang et~al.(2023)Jiang, Lu, Zhang, Ma, Han, Lyu, Li, and Chen]{jiang2023rtmpose}
Tao Jiang, Peng Lu, Li Zhang, Ningsheng Ma, Rui Han, Chengqi Lyu, Yining Li, and Kai Chen.
\newblock Rtmpose: Real-time multi-person pose estimation based on mmpose.
\newblock \emph{arXiv preprint arXiv:2303.07399}, 2023.

\bibitem[Jiang et~al.(2025{\natexlab{b}})Jiang, Han, Mao, Zhang, Pan, and Liu]{25iccv_vace}
Zeyinzi Jiang, Zhen Han, Chaojie Mao, Jingfeng Zhang, Yulin Pan, and Yu Liu.
\newblock Vace: All-in-one video creation and editing.
\newblock In \emph{Proceedings of the IEEE/CVF International Conference on Computer Vision}, pages 17191--17202, 2025{\natexlab{b}}.

\bibitem[Ju et~al.(2025)Ju, Ye, Liu, Wang, Wang, Wan, Zhang, Gai, and Xu]{ju2025fulldit}
Xuan Ju, Weicai Ye, Quande Liu, Qiulin Wang, Xintao Wang, Pengfei Wan, Di Zhang, Kun Gai, and Qiang Xu.
\newblock Fulldit: Multi-task video generative foundation model with full attention.
\newblock \emph{arXiv preprint arXiv:2503.19907}, 2025.

\bibitem[Kingma(2014)]{kingma2014adam}
Diederik~P Kingma.
\newblock Adam: A method for stochastic optimization.
\newblock \emph{arXiv preprint arXiv:1412.6980}, 2014.

\bibitem[Kong et~al.(2024)Kong, Tian, Zhang, Min, Dai, Zhou, Xiong, Li, Wu, Zhang, et~al.]{kong2024hunyuanvideo}
Weijie Kong, Qi Tian, Zijian Zhang, Rox Min, Zuozhuo Dai, Jin Zhou, Jiangfeng Xiong, Xin Li, Bo Wu, Jianwei Zhang, et~al.
\newblock Hunyuanvideo: A systematic framework for large video generative models.
\newblock \emph{arXiv preprint arXiv:2412.03603}, 2024.

\bibitem[Kong et~al.(2025)Kong, Gao, Zhang, Kang, Wei, Cai, Chen, and Luo]{kong2025let}
Zhe Kong, Feng Gao, Yong Zhang, Zhuoliang Kang, Xiaoming Wei, Xunliang Cai, Guanying Chen, and Wenhan Luo.
\newblock Let them talk: Audio-driven multi-person conversational video generation.
\newblock \emph{arXiv preprint arXiv:2505.22647}, 2025.

\bibitem[Kuaishou(2024.06)]{kuaishou2024kling}
Kuaishou.
\newblock Kling ai.
\newblock \emph{https://klingai.kuaishou.com/}, 2024.06.

\bibitem[Li et~al.(2024)Li, Li, Yang, Cao, Zhu, Cheng, and Chen]{li2024dispose}
Hongxiang Li, Yaowei Li, Yuhang Yang, Junjie Cao, Zhihong Zhu, Xuxin Cheng, and Long Chen.
\newblock Dispose: Disentangling pose guidance for controllable human image animation.
\newblock \emph{arXiv preprint arXiv:2412.09349}, 2024.

\bibitem[Li et~al.(2025)Li, Li, Lin, Niu, Yang, Huang, Cai, Jiang, Hu, and Chen]{li2025gir}
Hongxiang Li, Yaowei Li, Bin Lin, Yuwei Niu, Yuhang Yang, Xiaoshuang Huang, Jiayin Cai, Xiaolong Jiang, Yao Hu, and Long Chen.
\newblock Gir-bench: Versatile benchmark for generating images with reasoning.
\newblock \emph{arXiv preprint arXiv:2510.11026}, 2025.

\bibitem[Liang et~al.(2025)Liang, Yu, Zhou, Hu, Wang, Chen, Lin, Zhou, Li, Lu, et~al.]{liang2025omniv2v}
Sen Liang, Zhentao Yu, Zhengguang Zhou, Teng Hu, Hongmei Wang, Yi Chen, Qin Lin, Yuan Zhou, Xin Li, Qinglin Lu, et~al.
\newblock Omniv2v: Versatile video generation and editing via dynamic content manipulation.
\newblock \emph{arXiv preprint arXiv:2506.01801}, 2025.

\bibitem[Lin et~al.(2024)Lin, Jiang, Liang, Zhong, Yang, and Zheng]{lin2024cyberhost}
Gaojie Lin, Jianwen Jiang, Chao Liang, Tianyun Zhong, Jiaqi Yang, and Yanbo Zheng.
\newblock Cyberhost: Taming audio-driven avatar diffusion model with region codebook attention.
\newblock \emph{arXiv preprint arXiv:2409.01876}, 2024.

\bibitem[Lin et~al.(2025)Lin, Jiang, Yang, Zheng, and Liang]{lin2025omnihuman1}
Gaojie Lin, Jianwen Jiang, Jiaqi Yang, Zerong Zheng, and Chao Liang.
\newblock Omnihuman-1: Rethinking the scaling-up of one-stage conditioned human animation models.
\newblock \emph{arXiv preprint arXiv:2502.01061}, 2025.

\bibitem[Ling et~al.(2025)Ling, Zhu, Wu, Li, Feng, Yang, Hao, Zhu, Wu, and Chu]{ling2025vmbench}
Xinran Ling, Chen Zhu, Meiqi Wu, Hangyu Li, Xiaokun Feng, Cundian Yang, Aiming Hao, Jiashu Zhu, Jiahong Wu, and Xiangxiang Chu.
\newblock Vmbench: A benchmark for perception-aligned video motion generation.
\newblock In \emph{Proceedings of the IEEE/CVF International Conference on Computer Vision}, pages 13087--13098, 2025.

\bibitem[Lipman et~al.(2022)Lipman, Chen, Ben-Hamu, Nickel, and Le]{lipman2022flow}
Yaron Lipman, Ricky~TQ Chen, Heli Ben-Hamu, Maximilian Nickel, and Matt Le.
\newblock Flow matching for generative modeling.
\newblock \emph{arXiv preprint arXiv:2210.02747}, 2022.

\bibitem[Liu et~al.(2024)Liu, Zhai, Yang, Luo, Liang, Cao, and Zha]{liu2024grounding}
Cuiyu Liu, Wei Zhai, Yuhang Yang, Hongchen Luo, Sen Liang, Yang Cao, and Zheng-Jun Zha.
\newblock Grounding 3d scene affordance from egocentric interactions.
\newblock \emph{arXiv preprint arXiv:2409.19650}, 2024.

\bibitem[Liu et~al.(2025{\natexlab{a}})Liu, Ma, Li, Chen, Liu, Li, Zhou, He, and Wu]{liu2025phantom}
Lijie Liu, Tianxiang Ma, Bingchuan Li, Zhuowei Chen, Jiawei Liu, Gen Li, Siyu Zhou, Qian He, and Xinglong Wu.
\newblock Phantom: Subject-consistent video generation via cross-modal alignment.
\newblock \emph{arXiv preprint arXiv:2502.11079}, 2025{\natexlab{a}}.

\bibitem[Liu et~al.(2023{\natexlab{a}})Liu, Zeng, Ren, Li, Zhang, Yang, Li, Yang, Su, Zhu, et~al.]{liu2023grounding}
Shilong Liu, Zhaoyang Zeng, Tianhe Ren, Feng Li, Hao Zhang, Jie Yang, Chunyuan Li, Jianwei Yang, Hang Su, Jun Zhu, et~al.
\newblock Grounding dino: Marrying dino with grounded pre-training for open-set object detection.
\newblock \emph{arXiv preprint arXiv:2303.05499}, 2023{\natexlab{a}}.

\bibitem[Liu et~al.(2023{\natexlab{b}})Liu, Feng, Zhu, Zhang, Zheng, Liu, Zhao, Zhou, and Cao]{liu2023cones}
Zhiheng Liu, Ruili Feng, Kai Zhu, Yifei Zhang, Kecheng Zheng, Yu Liu, Deli Zhao, Jingren Zhou, and Yang Cao.
\newblock Cones: Concept neurons in diffusion models for customized generation.
\newblock \emph{arXiv preprint arXiv:2303.05125}, 2023{\natexlab{b}}.

\bibitem[Liu et~al.(2023{\natexlab{c}})Liu, Zhang, Shen, Zheng, Zhu, Feng, Liu, Zhao, Zhou, and Cao]{liu2023customizable}
Zhiheng Liu, Yifei Zhang, Yujun Shen, Kecheng Zheng, Kai Zhu, Ruili Feng, Yu Liu, Deli Zhao, Jingren Zhou, and Yang Cao.
\newblock Customizable image synthesis with multiple subjects.
\newblock \emph{Advances in neural information processing systems}, 36:\penalty0 57500--57519, 2023{\natexlab{c}}.

\bibitem[Liu et~al.(2025{\natexlab{b}})Liu, Cheng, Chen, Xiao, Ouyang, Zhu, Liu, Shen, Chen, and Luo]{liu2025manganinja}
Zhiheng Liu, Ka~Leong Cheng, Xi Chen, Jie Xiao, Hao Ouyang, Kai Zhu, Yu Liu, Yujun Shen, Qifeng Chen, and Ping Luo.
\newblock Manganinja: Line art colorization with precise reference following.
\newblock In \emph{Proceedings of the Computer Vision and Pattern Recognition Conference}, pages 5666--5677, 2025{\natexlab{b}}.

\bibitem[Liu et~al.(2025{\natexlab{c}})Liu, Deng, Chen, Wang, Guo, Han, Xue, Chen, Luo, and Yang]{liu2025worldweaver}
Zhiheng Liu, Xueqing Deng, Shoufa Chen, Angtian Wang, Qiushan Guo, Mingfei Han, Zeyue Xue, Mengzhao Chen, Ping Luo, and Linjie Yang.
\newblock Worldweaver: Generating long-horizon video worlds via rich perception.
\newblock \emph{arXiv preprint arXiv:2508.15720}, 2025{\natexlab{c}}.

\bibitem[Liu et~al.(2025{\natexlab{d}})Liu, Ren, Liu, Zhou, Chen, Qiu, Huang, An, Yang, Patel, et~al.]{liu2025tuna}
Zhiheng Liu, Weiming Ren, Haozhe Liu, Zijian Zhou, Shoufa Chen, Haonan Qiu, Xiaoke Huang, Zhaochong An, Fanny Yang, Aditya Patel, et~al.
\newblock Tuna: Taming unified visual representations for native unified multimodal models.
\newblock \emph{arXiv preprint arXiv:2512.02014}, 2025{\natexlab{d}}.

\bibitem[Loper et~al.(2015)Loper, Mahmood, Romero, Pons-Moll, and Black]{15tog_smpl}
Matthew Loper, Naureen Mahmood, Javier Romero, Gerard Pons-Moll, and Michael~J. Black.
\newblock {SMPL}: A skinned multi-person linear model.
\newblock \emph{ACM Trans. Graph.}, 2015.

\bibitem[Meng et~al.(2025)Meng, Wang, Wu, Zheng, Li, and Ma]{meng2025echomimicv3}
Rang Meng, Yan Wang, Weipeng Wu, Ruobing Zheng, Yuming Li, and Chenguang Ma.
\newblock Echomimicv3: 1.3b parameters are all you need for unified multi-modal and multi-task human animation, 2025.

\bibitem[Polyak et~al.(2024)Polyak, Zohar, Brown, Tjandra, Sinha, Lee, Vyas, Shi, Ma, Chuang, et~al.]{polyak2024movie}
Adam Polyak, Amit Zohar, Andrew Brown, Andros Tjandra, Animesh Sinha, Ann Lee, Apoorv Vyas, Bowen Shi, Chih-Yao Ma, Ching-Yao Chuang, et~al.
\newblock Movie gen: A cast of media foundation models.
\newblock \emph{arXiv preprint arXiv:2410.13720}, 2024.

\bibitem[Qiu et~al.(2025)Qiu, Gu, Li, Zuo, Shen, Zhang, Qiu, Yuan, Chen, Dong, et~al.]{qiu2025lhm}
Lingteng Qiu, Xiaodong Gu, Peihao Li, Qi Zuo, Weichao Shen, Junfei Zhang, Kejie Qiu, Weihao Yuan, Guanying Chen, Zilong Dong, et~al.
\newblock Lhm: Large animatable human reconstruction model from a single image in seconds.
\newblock \emph{arXiv preprint arXiv:2503.10625}, 2025.

\bibitem[Radford et~al.(2021)Radford, Kim, Hallacy, Ramesh, Goh, Agarwal, Sastry, Askell, Mishkin, Clark, et~al.]{radford2021learning}
Alec Radford, Jong~Wook Kim, Chris Hallacy, Aditya Ramesh, Gabriel Goh, Sandhini Agarwal, Girish Sastry, Amanda Askell, Pamela Mishkin, Jack Clark, et~al.
\newblock Learning transferable visual models from natural language supervision.
\newblock In \emph{International conference on machine learning}, pages 8748--8763. PmLR, 2021.

\bibitem[Savchenko(2023)]{savchenko2023facial}
Andrey Savchenko.
\newblock Facial expression recognition with adaptive frame rate based on multiple testing correction.
\newblock In \emph{Proceedings of the 40th International Conference on Machine Learning (ICML)}, pages 30119--30129. PMLR, 2023.

\bibitem[Savchenko et~al.(2022)Savchenko, Savchenko, and Makarov]{savchenko2022classifying}
Andrey~V Savchenko, Lyudmila~V Savchenko, and Ilya Makarov.
\newblock Classifying emotions and engagement in online learning based on a single facial expression recognition neural network.
\newblock \emph{IEEE Transactions on Affective Computing}, 2022.

\bibitem[Seedream et~al.(2025)Seedream, Chen, Gao, Gong, Guo, Guo, Guo, Hou, Huang, Huang, et~al.]{seedream2025seedream}
Team Seedream, Yunpeng Chen, Yu Gao, Lixue Gong, Meng Guo, Qiushan Guo, Zhiyao Guo, Xiaoxia Hou, Weilin Huang, Yixuan Huang, et~al.
\newblock Seedream 4.0: Toward next-generation multimodal image generation.
\newblock \emph{arXiv preprint arXiv:2509.20427}, 2025.

\bibitem[Shao et~al.(2025)Shao, Zhai, Yang, Luo, Cao, and Zha]{shao2025great}
Yawen Shao, Wei Zhai, Yuhang Yang, Hongchen Luo, Yang Cao, and Zheng-Jun Zha.
\newblock Great: Geometry-intention collaborative inference for open-vocabulary 3d object affordance grounding.
\newblock In \emph{Proceedings of the Computer Vision and Pattern Recognition Conference}, pages 17326--17336, 2025.

\bibitem[Shen et~al.(2024)Shen, Pi, Xia, Cen, Peng, Hu, Bao, Hu, and Zhou]{24siga_gvhmr}
Zehong Shen, Huaijin Pi, Yan Xia, Zhi Cen, Sida Peng, Zechen Hu, Hujun Bao, Ruizhen Hu, and Xiaowei Zhou.
\newblock World-grounded human motion recovery via gravity-view coordinates.
\newblock In \emph{SIGGRAPH Asia Conference Proceedings}, 2024.

\bibitem[Sim{\'e}oni et~al.(2025)Sim{\'e}oni, Vo, Seitzer, Baldassarre, Oquab, Jose, Khalidov, Szafraniec, Yi, Ramamonjisoa, Massa, Haziza, Wehrstedt, Wang, Darcet, Moutakanni, Sentana, Roberts, Vedaldi, Tolan, Brandt, Couprie, Mairal, J{\'e}gou, Labatut, and Bojanowski]{simeoni2025dinov3}
Oriane Sim{\'e}oni, Huy~V. Vo, Maximilian Seitzer, Federico Baldassarre, Maxime Oquab, Cijo Jose, Vasil Khalidov, Marc Szafraniec, Seungeun Yi, Micha{\"e}l Ramamonjisoa, Francisco Massa, Daniel Haziza, Luca Wehrstedt, Jianyuan Wang, Timoth{\'e}e Darcet, Th{\'e}o Moutakanni, Leonel Sentana, Claire Roberts, Andrea Vedaldi, Jamie Tolan, John Brandt, Camille Couprie, Julien Mairal, Herv{\'e} J{\'e}gou, Patrick Labatut, and Piotr Bojanowski.
\newblock {DINOv3}, 2025.

\bibitem[Su et~al.(2024)Su, Ahmed, Lu, Pan, Bo, and Liu]{su2024roformer}
Jianlin Su, Murtadha Ahmed, Yu Lu, Shengfeng Pan, Wen Bo, and Yunfeng Liu.
\newblock Roformer: Enhanced transformer with rotary position embedding.
\newblock \emph{Neurocomputing}, 568:\penalty0 127063, 2024.

\bibitem[Teed and Deng(2021)]{teed2021droid}
Zachary Teed and Jia Deng.
\newblock Droid-slam: Deep visual slam for monocular, stereo, and rgb-d cameras.
\newblock \emph{Advances in neural information processing systems}, 34:\penalty0 16558--16569, 2021.

\bibitem[Wan et~al.(2025)Wan, Wang, Ai, Wen, Mao, Xie, Chen, Yu, Zhao, Yang, Zeng, Wang, Zhang, Zhou, Wang, Chen, Zhu, Zhao, Yan, Huang, Feng, Zhang, Li, Wu, Chu, Feng, Zhang, Sun, Fang, Wang, Gui, Weng, Shen, Lin, Wang, Wang, Zhou, Wang, Shen, Yu, Shi, Huang, Xu, Kou, Lv, Li, Liu, Wang, Zhang, Huang, Li, Wu, Liu, Pan, Zheng, Hong, Shi, Feng, Jiang, Han, Wu, and Liu]{wan2025}
Team Wan, Ang Wang, Baole Ai, Bin Wen, Chaojie Mao, Chen-Wei Xie, Di Chen, Feiwu Yu, Haiming Zhao, Jianxiao Yang, Jianyuan Zeng, Jiayu Wang, Jingfeng Zhang, Jingren Zhou, Jinkai Wang, Jixuan Chen, Kai Zhu, Kang Zhao, Keyu Yan, Lianghua Huang, Mengyang Feng, Ningyi Zhang, Pandeng Li, Pingyu Wu, Ruihang Chu, Ruili Feng, Shiwei Zhang, Siyang Sun, Tao Fang, Tianxing Wang, Tianyi Gui, Tingyu Weng, Tong Shen, Wei Lin, Wei Wang, Wei Wang, Wenmeng Zhou, Wente Wang, Wenting Shen, Wenyuan Yu, Xianzhong Shi, Xiaoming Huang, Xin Xu, Yan Kou, Yangyu Lv, Yifei Li, Yijing Liu, Yiming Wang, Yingya Zhang, Yitong Huang, Yong Li, You Wu, Yu Liu, Yulin Pan, Yun Zheng, Yuntao Hong, Yupeng Shi, Yutong Feng, Zeyinzi Jiang, Zhen Han, Zhi-Fan Wu, and Ziyu Liu.
\newblock Wan: Open and advanced large-scale video generative models.
\newblock \emph{arXiv preprint arXiv:2503.20314}, 2025.

\bibitem[Wang et~al.(2024)Wang, Yeh, and Mark~Liao]{wang2024yolov9}
Chien-Yao Wang, I-Hau Yeh, and Hong-Yuan Mark~Liao.
\newblock Yolov9: Learning what you want to learn using programmable gradient information.
\newblock In \emph{European conference on computer vision}, pages 1--21. Springer, 2024.

\bibitem[Wang et~al.(2025{\natexlab{a}})Wang, Wang, Jiang, Fan, Zhang, Qi, Zhao, and Xu]{wang2025fantasytalking}
Mengchao Wang, Qiang Wang, Fan Jiang, Yaqi Fan, Yunpeng Zhang, Yonggang Qi, Kun Zhao, and Mu Xu.
\newblock Fantasytalking: Realistic talking portrait generation via coherent motion synthesis.
\newblock \emph{arXiv preprint arXiv:2504.04842}, 2025{\natexlab{a}}.

\bibitem[Wang et~al.(2025{\natexlab{b}})Wang, Hu, Huang, Su, Yi, and Ma]{wang2025poseanything}
Ruiyan Wang, Teng Hu, Kaihui Huang, Zihan Su, Ran Yi, and Lizhuang Ma.
\newblock Poseanything: Universal pose-guided video generation with part-aware temporal coherence.
\newblock \emph{arXiv preprint arXiv:2512.13465}, 2025{\natexlab{b}}.

\bibitem[Wang et~al.(2025{\natexlab{c}})Wang, Wang, Zhang, Fan, Wu, Xue, and Liu]{wang2025timotion}
Yabiao Wang, Shuo Wang, Jiangning Zhang, Ke Fan, Jiafu Wu, Zhucun Xue, and Yong Liu.
\newblock Timotion: Temporal and interactive framework for efficient human-human motion generation.
\newblock In \emph{Proceedings of the Computer Vision and Pattern Recognition Conference}, pages 7169--7178, 2025{\natexlab{c}}.

\bibitem[Wei et~al.(2025{\natexlab{a}})Wei, Liu, Ye, Wang, Wang, Wan, Gai, and Chen]{25_univideo}
Cong Wei, Quande Liu, Zixuan Ye, Qiulin Wang, Xintao Wang, Pengfei Wan, Kun Gai, and Wenhu Chen.
\newblock Univideo: Unified understanding, generation, and editing for videos.
\newblock \emph{arXiv preprint arXiv:2510.08377}, 2025{\natexlab{a}}.

\bibitem[Wei et~al.(2025{\natexlab{b}})Wei, Sun, Ma, Hou, Juefei-Xu, He, Dai, Zhang, Li, Hou, et~al.]{wei2025mocha}
Cong Wei, Bo Sun, Haoyu Ma, Ji Hou, Felix Juefei-Xu, Zecheng He, Xiaoliang Dai, Luxin Zhang, Kunpeng Li, Tingbo Hou, et~al.
\newblock Mocha: Towards movie-grade talking character synthesis.
\newblock \emph{arXiv preprint arXiv:2503.23307}, 2025{\natexlab{b}}.

\bibitem[Wiedemer et~al.(2025)Wiedemer, Li, Vicol, Gu, Matarese, Swersky, Kim, Jaini, and Geirhos]{wiedemer2025video}
Thadd{\"a}us Wiedemer, Yuxuan Li, Paul Vicol, Shixiang~Shane Gu, Nick Matarese, Kevin Swersky, Been Kim, Priyank Jaini, and Robert Geirhos.
\newblock Video models are zero-shot learners and reasoners.
\newblock \emph{arXiv preprint arXiv:2509.20328}, 2025.

\bibitem[Wu et~al.(2025)Wu, Li, Zhou, Lin, Gao, Yan, ming Yin, Bai, Xu, Chen, Chen, Tang, Zhang, Wang, Yang, Yu, Cheng, Liu, Li, Zhang, Meng, Wei, Ni, Chen, Cao, Peng, Qu, Wu, Wang, Yu, Wen, Feng, Xu, Wang, Zhang, Zhu, Wu, Cai, and Liu]{wu2025qwenimagetechnicalreport}
Chenfei Wu, Jiahao Li, Jingren Zhou, Junyang Lin, Kaiyuan Gao, Kun Yan, Sheng ming Yin, Shuai Bai, Xiao Xu, Yilei Chen, Yuxiang Chen, Zecheng Tang, Zekai Zhang, Zhengyi Wang, An Yang, Bowen Yu, Chen Cheng, Dayiheng Liu, Deqing Li, Hang Zhang, Hao Meng, Hu Wei, Jingyuan Ni, Kai Chen, Kuan Cao, Liang Peng, Lin Qu, Minggang Wu, Peng Wang, Shuting Yu, Tingkun Wen, Wensen Feng, Xiaoxiao Xu, Yi Wang, Yichang Zhang, Yongqiang Zhu, Yujia Wu, Yuxuan Cai, and Zenan Liu.
\newblock Qwen-image technical report, 2025.

\bibitem[Wu et~al.(2023)Wu, Zhang, Zhang, Chen, Liao, Li, Gao, Wang, Zhang, Sun, et~al.]{wu2023q}
Haoning Wu, Zicheng Zhang, Weixia Zhang, Chaofeng Chen, Liang Liao, Chunyi Li, Yixuan Gao, Annan Wang, Erli Zhang, Wenxiu Sun, et~al.
\newblock Q-align: Teaching lmms for visual scoring via discrete text-defined levels.
\newblock \emph{arXiv preprint arXiv:2312.17090}, 2023.

\bibitem[Xiao et~al.(2025)Xiao, Lu, Pi, Fan, Pan, Zhou, Feng, Zhou, Peng, and Wang]{25iccv_motionstreamer}
Lixing Xiao, Shunlin Lu, Huaijin Pi, Ke Fan, Liang Pan, Yueer Zhou, Ziyong Feng, Xiaowei Zhou, Sida Peng, and Jingbo Wang.
\newblock Motionstreamer: Streaming motion generation via diffusion-based autoregressive model in causal latent space.
\newblock In \emph{Proceedings of the IEEE/CVF International Conference on Computer Vision (ICCV)}, pages 10086--10096, 2025.

\bibitem[Xu et~al.(2024)Xu, Zhang, Liew, Yan, Liu, Zhang, Feng, and Shou]{24cvpr_magicanimate}
Zhongcong Xu, Jianfeng Zhang, Jun~Hao Liew, Hanshu Yan, Jia-Wei Liu, Chenxu Zhang, Jiashi Feng, and Mike~Zheng Shou.
\newblock Magicanimate: Temporally consistent human image animation using diffusion model.
\newblock 2024.

\bibitem[Yang et~al.(2025{\natexlab{a}})Yang, Kong, Gao, Cheng, Liu, Zhang, Kang, Luo, Cai, He, and Wei]{yang2025infinitetalkaudiodrivenvideogeneration}
Shaoshu Yang, Zhe Kong, Feng Gao, Meng Cheng, Xiangyu Liu, Yong Zhang, Zhuoliang Kang, Wenhan Luo, Xunliang Cai, Ran He, and Xiaoming Wei.
\newblock Infinitetalk: Audio-driven video generation for sparse-frame video dubbing, 2025{\natexlab{a}}.

\bibitem[Yang et~al.(2024{\natexlab{a}})Yang, Zhai, Luo, Cao, and Zha]{yang2024lemon}
Yuhang Yang, Wei Zhai, Hongchen Luo, Yang Cao, and Zheng-Jun Zha.
\newblock Lemon: Learning 3d human-object interaction relation from 2d images.
\newblock In \emph{Proceedings of the IEEE/CVF Conference on Computer Vision and Pattern Recognition}, pages 16284--16295, 2024{\natexlab{a}}.

\bibitem[Yang et~al.(2024{\natexlab{b}})Yang, Zhai, Wang, Yu, Cao, and Zha]{yang2024egochoir}
Yuhang Yang, Wei Zhai, Chengfeng Wang, Chengjun Yu, Yang Cao, and Zheng-Jun Zha.
\newblock Egochoir: Capturing 3d human-object interaction regions from egocentric views.
\newblock \emph{Advances in Neural Information Processing Systems}, 37:\penalty0 54529--54557, 2024{\natexlab{b}}.

\bibitem[Yang et~al.(2025{\natexlab{b}})Yang, Fan, Sun, Li, Zeng, Han, Zhai, Liu, Cao, and Zha]{yang2025videogen}
Yuhang Yang, Ke Fan, Shangkun Sun, Hongxiang Li, Ailing Zeng, FeiLin Han, Wei Zhai, Wei Liu, Yang Cao, and Zheng-Jun Zha.
\newblock Videogen-eval: Agent-based system for video generation evaluation.
\newblock \emph{arXiv preprint arXiv:2503.23452}, 2025{\natexlab{b}}.

\bibitem[Yang et~al.(2025{\natexlab{c}})Yang, Liu, Lu, Zhao, Wu, Zhai, Yi, Cao, Ma, Zha, et~al.]{yang2025sigman}
Yuhang Yang, Fengqi Liu, Yixing Lu, Qin Zhao, Pingyu Wu, Wei Zhai, Ran Yi, Yang Cao, Lizhuang Ma, Zheng-Jun Zha, et~al.
\newblock Sigman: Scaling 3d human gaussian generation with millions of assets.
\newblock In \emph{Proceedings of the IEEE/CVF International Conference on Computer Vision}, pages 5122--5133, 2025{\natexlab{c}}.

\bibitem[Yang et~al.(2023)Yang, Zeng, Yuan, and Li]{yang2023effective}
Zhendong Yang, Ailing Zeng, Chun Yuan, and Yu Li.
\newblock Effective whole-body pose estimation with two-stages distillation.
\newblock In \emph{Proceedings of the IEEE/CVF International Conference on Computer Vision}, pages 4210--4220, 2023.

\bibitem[Yu et~al.(2025)Yu, Zhai, Yang, Cao, and Zha]{yu2025hero}
Chengjun Yu, Wei Zhai, Yuhang Yang, Yang Cao, and Zheng-Jun Zha.
\newblock Hero: Human reaction generation from videos.
\newblock In \emph{Proceedings of the IEEE/CVF International Conference on Computer Vision}, pages 10262--10274, 2025.

\bibitem[Zeng et~al.(2024)Zeng, Yang, Chen, and Liu]{zeng2024dawn}
Ailing Zeng, Yuhang Yang, Weidong Chen, and Wei Liu.
\newblock The dawn of video generation: Preliminary explorations with sora-like models.
\newblock \emph{arXiv preprint arXiv:2410.05227}, 2024.

\bibitem[Zhang et~al.(2024)Zhang, Tian, Huang, Qiao, and Liu]{zhang2024evaluationagent}
Fan Zhang, Shulin Tian, Ziqi Huang, Yu Qiao, and Ziwei Liu.
\newblock Evaluation agent: Efficient and promptable evaluation framework for visual generative models.
\newblock In \emph{Annual Meeting of the Association for Computational Linguistics (ACL), 2025}, 2024.

\bibitem[Zhang and Agrawala(2025)]{zhang2025packing}
Lvmin Zhang and Maneesh Agrawala.
\newblock Packing input frame context in next-frame prediction models for video generation.
\newblock \emph{arXiv preprint arXiv:2504.12626}, 2025.

\bibitem[Zhu et~al.(2022)Zhu, Wu, Zhu, Jiang, Tang, Zhang, Liu, and Loy]{zhu2022celebvhq}
Hao Zhu, Wayne Wu, Wentao Zhu, Liming Jiang, Siwei Tang, Li Zhang, Ziwei Liu, and Chen~Change Loy.
\newblock {CelebV-HQ}: A large-scale video facial attributes dataset.
\newblock In \emph{ECCV}, 2022.

\end{thebibliography}
}

\section{Appendix}

\subsection{Details of Data Pipeline}
Here, we provide additional details on the pipeline for extracting anchor frames. For the global anchor, since we can control how each $5$s training clip is segmented from a long video, we know the exact start and end timestamps of each clip. This allows us to extract global frames both inside and outside the training clip.

Regarding the viewpoint anchor, we leverage the world-grounded human motion recovery method GVHMR \cite{24siga_gvhmr}. It represents human motion using the 3D body model SMPL \cite{15tog_smpl} (a widely used parameterized human model \cite{liu2023grounding,yang2024lemon,yang2024egochoir,fan2024freemotion,25iccv_motionstreamer,wang2025timotion,fan2025go,fan2024textual}) and recovers human motion directly from videos. Specifically, it estimates the facing direction of the person, which can be represented by the Z-axis orientation of the SMPL root joint. In addition, such methods typically use SLAM \cite{teed2021droid} to estimate the camera pose. By computing the angle $\theta$ between the camera orientation and the human orientation, and determining a threshold based on empirical observations across multiple test cases, we can distinguish frames in which the person is facing front, backward, left, or right. These categorized frames are then extracted as viewpoint-anchor frames.

For expression anchors, we adopt a hybrid extraction strategy. We first use EmotiEffLib \cite{savchenko2023facial,savchenko2022classifying} to scan video clips and identify frames containing any of the predefined eight expression categories, selecting only clips that include at least two distinct expressions. Although this method effectively identifies candidate clips, the expression category assigned to each extracted anchor frame may still be ambiguous. To address this issue, we further use an MLLM (\eg, Gemini \cite{comanici2025gemini}) to verify whether the predicted expression matches the semantic content of each anchor frame. If the prediction is incorrect, we correct the frame to the appropriate category through the MLLM; if it does not belong to any defined category, we discard it. This refinement process increases the accuracy from $66\%$ to $82\%$.

\subsection{Training and Inference Details}
\label{sec:train_detail}
Here we provide additional training details for content anchors. First, video clips that simultaneously contain different expressions and viewpoints are relatively scarce. Thanks to our proposed ``RoPE as Weak Condition'', which binds each type of content anchor to a fixed position, the model can be trained in a mixed manner to learn how to extract information from different types of anchors. Specifically, we first balance the number of anchors across all categories, with the final ratio of viewpoint anchors is shown in Tab. \ref{tab:view_anchor}. The ratio of expression anchors is reported in Tab. \ref{tab:expression_anchor}. Then, we ensure that training batches contain a balanced mixture of expression-anchor frames and viewpoint-anchor frames, and training with a batch size of $256$ for $3000$ steps. Afterward, we perform a fine-tuning stage ($200$ steps) on a small subset of data that contains both types of anchors, allowing the model to adapt to scenarios where multiple anchor types appear simultaneously. Note that the global anchor remains throughout the entire training process.

Regarding the linear blend in continuation generation, let the last $n$ frames of the previous chunk be $x_1$, and the first $n$ frames of the next be $x_2$, blended as: $w*x_1+(1-w)*x_2$, $n=4$, $w=[1,0.67,0.33,0]$, applied at the clean latent level.
\begin{table}[t]
\centering
  \renewcommand{\arraystretch}{1.}
  \renewcommand{\tabcolsep}{3.5 pt}
  \caption{Ratios of different categories in viewpoint-anchor.}
\label{tab:view_anchor}
\begin{tabular}{c|cccc}
\toprule
\textbf{Viewpoints} & \textbf{Front} & \textbf{Back} & \textbf{Left} & \textbf{Right} \\ \midrule
\textbf{Ratio}      & 29.6$\%$              & 17.3$\%$             & 27.9$\%$             & 25.2$\%$              \\ \bottomrule
\end{tabular}
\end{table}

\begin{table}[t]
\centering
  \renewcommand{\arraystretch}{1.}
  \renewcommand{\tabcolsep}{3.5 pt}
  \caption{Ratios of different categories in expression-anchor.}
\label{tab:expression_anchor}
\begin{tabular}{c|cccc}
\toprule
\textbf{Expressions} & \textbf{Surprise} & \textbf{Angry}   & \textbf{Disgust} & \textbf{Fear} \\ \midrule
\textbf{Ratio}       & 15.0$\%$                 & 8.2$\%$                & 13.7$\%$                & 10.1$\%$             \\ \midrule
\textbf{Expressions} & \textbf{Contempt} & \textbf{Neutral} & \textbf{Happy}   & \textbf{Sad}  \\ \midrule
\textbf{Ratio}       & 13.6$\%$        & 8.1$\%$                & 14.6$\%$                & 16.7$\%$             \\ \bottomrule
\end{tabular}
\end{table}

\begin{figure*}[h]
    \centering
    \begin{overpic}[width=1.0\linewidth]{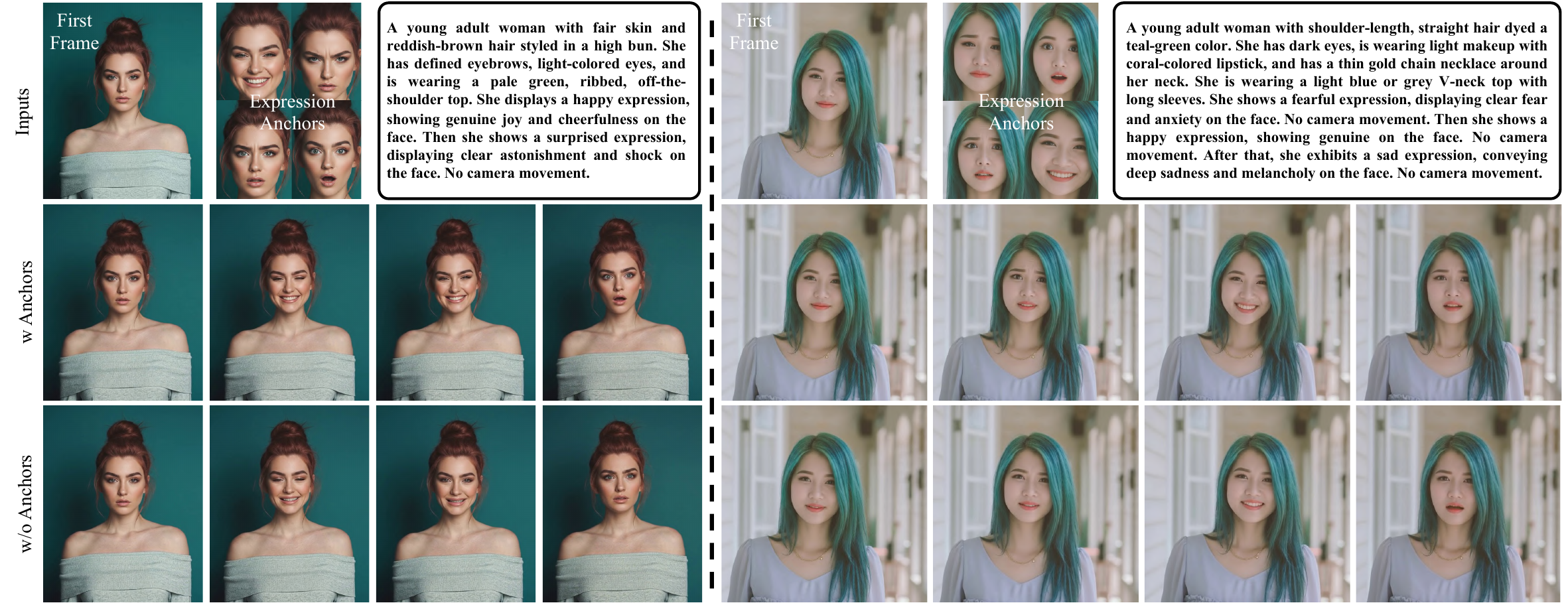}
    \end{overpic}
    \caption{Qualitative results of the expression anchors. The first row shows the inputs, the second row shows the results with the expression anchor, and the third row shows the results without the anchors.}
    \label{fig:supp_expression}
\end{figure*}

\subsection{Subject Evaluation Protocol}
Regarding the subject evaluation, we refer to certain dimensions of existing benchmarks \cite{zeng2024dawn, yang2025videogen,li2025gir,huang2023vbench} and make an A/B testing protocol. For each evaluation dimension, annotators are presented with a specific question to choose the better result, and they are also asked to provide the reason for the choice. Below, we list several example questions for reference.

\textbf{Questions for reference-based aspects:}
\begin{itemize}
    \item Between the two videos below, which one better preserves the multi-view appearance details shown in the reference images?
    \item Between the two videos below, which one better follows the prompt-specified expression, with facial details more closely matching the reference image?
    \item Between the two videos below, which one exhibits more natural overall motion when reference frames are provided?
\end{itemize}

\textbf{Questions for general aspects:}
\begin{itemize}
    \item Consider the following aspects for video color consistency: severity and frequency of overall color tone changes, background color shifts, and visual inconsistencies throughout the entire video. Which of the provided videos maintains the better overall color consistency?
    \item Consider the following aspects for movement naturalness: fluidity and continuity of actions and expressions, coordination between different body parts, naturalness of facial expression changes, realism and believability of movements, and overall naturalness of all movements throughout the entire video. Which videos have higher movement naturalness?
    \item Consider the following aspects for lip sync quality: severity and frequency of timing synchronization misalignment, pronunciation lip shape mismatch, and movement amplitude mismatch throughout the entire video. Which of the provided videos has higher lip sync quality?
\end{itemize}

\begin{figure}[t]
    \centering
    \begin{overpic}[width=1.0\linewidth]{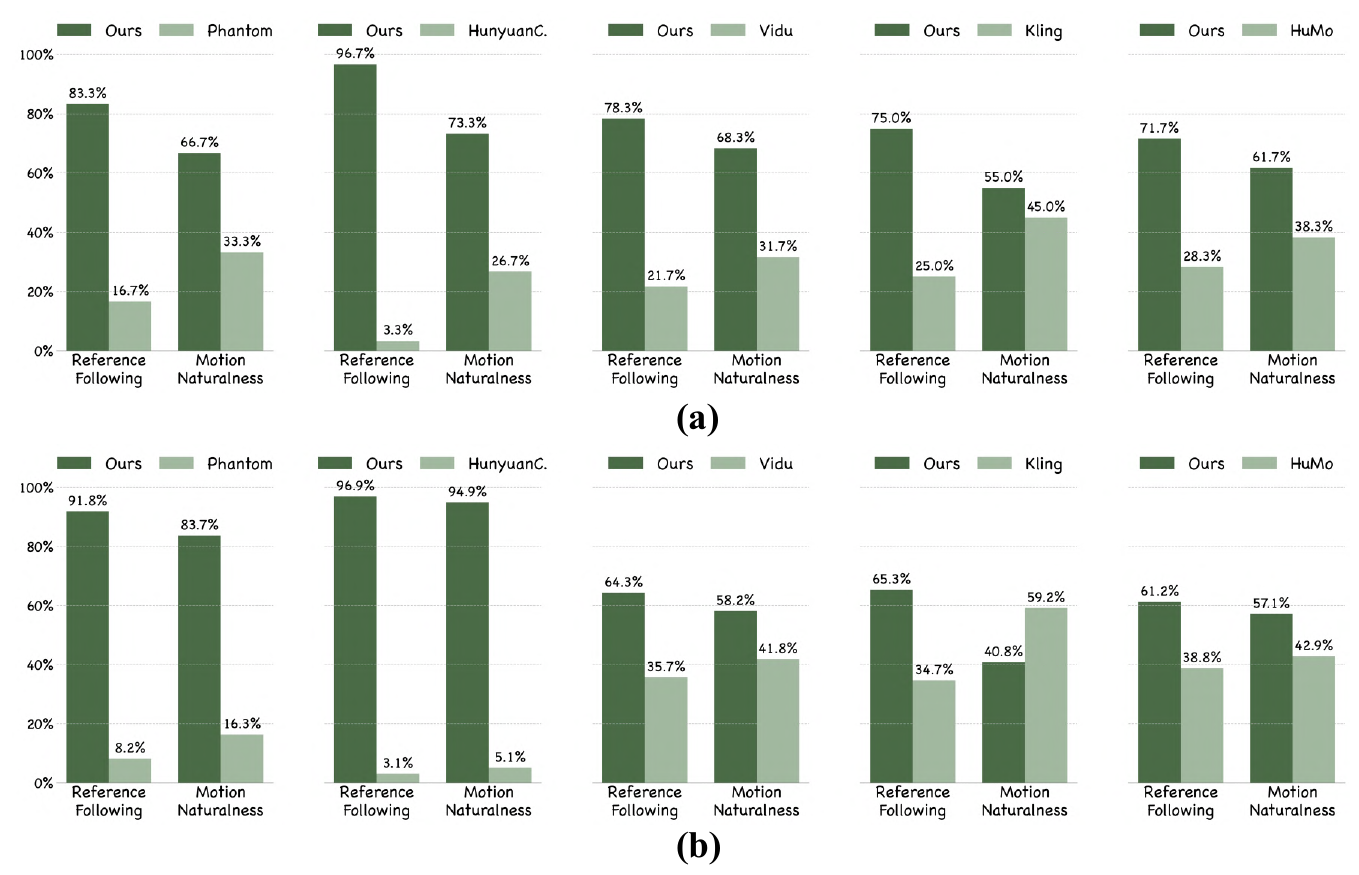}
    \end{overpic}
    \caption{Subject Evaluation of \textbf{(a)} expressive identity consistency, and \textbf{(b)} multi-view appearance consistency.}
    \label{fig:supp_user_ref}
\end{figure}

\subsection{More Experimental Results}
We provide more qualitative results illustrating the effect of the expression anchor. As shown in Fig. \ref{fig:supp_expression}, when the expression anchor is provided, the character produces more accurate facial expressions, and the facial details remain more consistent across different expressions.

\begin{figure}[t]
    \centering
    \begin{overpic}[width=1.0\linewidth]{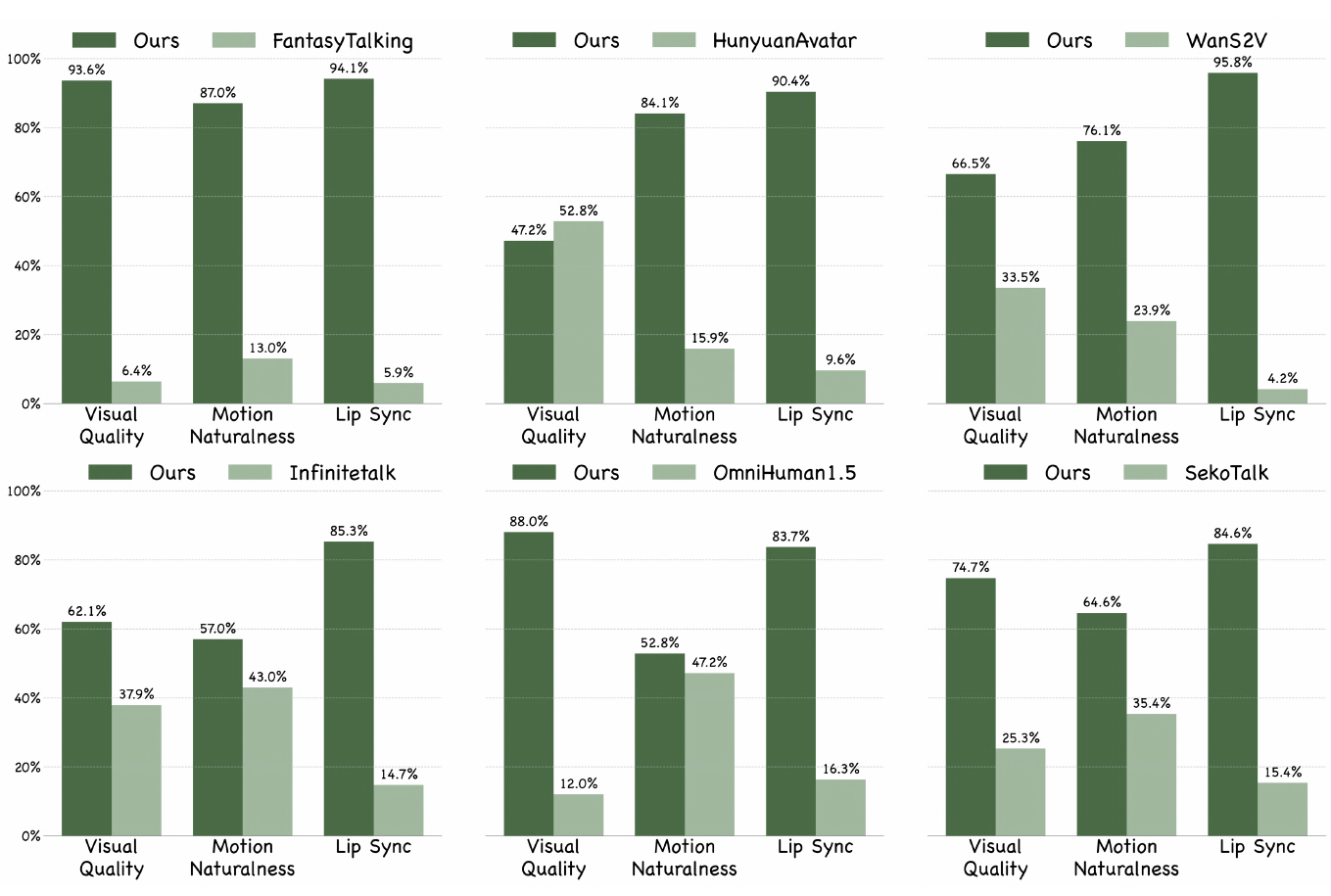}
    \end{overpic}
    \caption{Comparison of subject evaluation results with human-centric models across multiple evaluation dimensions.}
    \label{fig:supp_user_general}
\end{figure}

In the main paper, we present the overall subject evaluation results. Here, we provide more detailed results along individual dimensions: identity consistency across expressions is shown in Fig. \ref{fig:supp_user_ref} (a) and multi-view appearance consistency is shown in Fig. \ref{fig:supp_user_ref} (b), and comparisons with human-centric models are shown in Fig. \ref{fig:supp_user_general}.

In addition to the results on the combined testset from human-centric baselines reported in the main paper, we also provide comparisons against these baselines on public datasets such as CelebV-HQ \cite{zhu2022celebvhq}. We use $100$ cases in total, and the results are summarized in Tab. \ref{tab:celebv}. 

\begin{table}[t]
\centering
  \renewcommand{\arraystretch}{1.}
  \renewcommand{\tabcolsep}{1.5 pt}
  \caption{Quantitative comparison results on CelebV-HQ with the human-centric methods. Fantasy. denotes FantasyTalking, Huny.A. is HunyuanAvatar and Infinite. indicates InfiniteTalk.}
\label{tab:celebv}  
\begin{tabular}{c|cccccc}
\toprule
\textbf{Method}         & \multicolumn{1}{c}{\textbf{IQA}} $\uparrow$ & \textbf{AES} $\uparrow$ & \textbf{Sync-C} $\uparrow$ & \textbf{Sync-D} $\downarrow$ &  \textbf{FID/FVD} $\downarrow$ \\ \midrule
\textbf{Fantasy.} &           3.48       &      2.26        &              2.88   &    10.39          &   37.43 / 108.92          \\
\textbf{Huny.A.}  &           3.40            &          2.28              &        5.17         &       8.37       &       28.39 / 93.49        \\
\textbf{WanS2V}         &           3.56                  &       \textbf{2.33}       &         4.99        &        8.40      &     36.10 / 93.73         \\
\textbf{Infinite.}   &           3.47                     &    2.31          &       5.20          &       8.29       &      29.56 / \textbf{76.39}        \\
\textbf{Ours}           &           \textbf{3.59}                    &      \textbf{2.33}        &       \textbf{5.62}          &       \textbf{8.06}       &       \textbf{25.50} / 90.86       \\ \bottomrule
\end{tabular}
\end{table}%

We also compare the 5B and 17B models during training (10M data): the 5B model is weaker than the 17B across all dimensions. Regarding the data, we report the performance (17B) under 2M and 10M training clips, shown in Tab. \ref{tab:model_size}. This also aligns with the user study results.

\begin{table}[h]
\centering
\scriptsize
\renewcommand{\arraystretch}{1.3}
\renewcommand{\tabcolsep}{1.0 pt}
  \caption{Performance in different model sizes (a) and different data scales (b).}
\label{tab:model_size}
\begin{subtable}[t]{0.5\linewidth}
\begin{tabular}{c|cccc}
\Xhline{1pt}
Size  & IQA $\uparrow$ & AES $\uparrow$ & Sync-C $\uparrow$ & Arcface $\uparrow$\\ 
\Xhline{0.5pt}
5B  & 4.58 & 3.54 & 4.93 & 0.583\\
17B & 4.65 & 3.63 & 5.12 & 0.787\\
\Xhline{1pt}
\end{tabular}
\caption{}
\end{subtable}
\hfill
\begin{subtable}[t]{0.49\linewidth}
\begin{tabular}{c|cccc}
\Xhline{1pt}
Data  & IQA $\uparrow$ & AES $\uparrow$ & Sync-C $\uparrow$ & Arcface $\uparrow$\\ 
\Xhline{0.5pt}
2M  & 4.53 & 3.57 & 5.03 & 0.623\\
10M & 4.65 & 3.63 & 5.12 & 0.787\\
\Xhline{1pt}
\end{tabular}
\caption{}
\end{subtable}
\end{table}

\end{document}